\documentclass[]{xiaomi}

\usepackage[utf8]{inputenc} 
\usepackage[T1]{fontenc}    
\usepackage{hyperref}       
\usepackage{url}            
\usepackage{amsfonts}       
\usepackage{nicefrac}       
\usepackage{microtype}      
\usepackage{xcolor}         
\usepackage{wrapfig}
\usepackage{overpic}
\usepackage{rotating}   
\usepackage{graphicx}   
\usepackage{array}
\usepackage[table]{xcolor}
\usepackage{pifont}
\usepackage{textcomp}
\usepackage{booktabs}
\usepackage{multirow}
\usepackage{subcaption}   
\usepackage{amsmath}
\usepackage{marvosym}   
\usepackage{ragged2e}   
\usepackage{enumitem}

\usepackage{xspace}
\usepackage{siunitx} 
\definecolor{tblgray}{gray}{0.94}
\definecolor{pdmgray}{gray}{0.5}
\newcommand{\pdm}[1]{\textcolor{pdmgray}{#1}}
\newcommand{\pdmnote}{\textsuperscript{$\dagger$}}
\usepackage{wrapfig}  
\definecolor{dark-gray}{gray}{0.30}

\newcommand{\figref}[1]{Fig.~\ref{#1}}
\newcommand{\tabref}[1]{Tab.~\ref{#1}}

\newcommand{\name}{UniDriveVLA}

\newcommand{\pub}[1]{{\color{dark-gray}{\scriptsize{[{#1}]}}}}

\definecolor{tblgray}{gray}{0.94}

\title{\raisebox{-0.2\height}{\includegraphics[width=1.0cm,trim=10 10 10 10,clip]{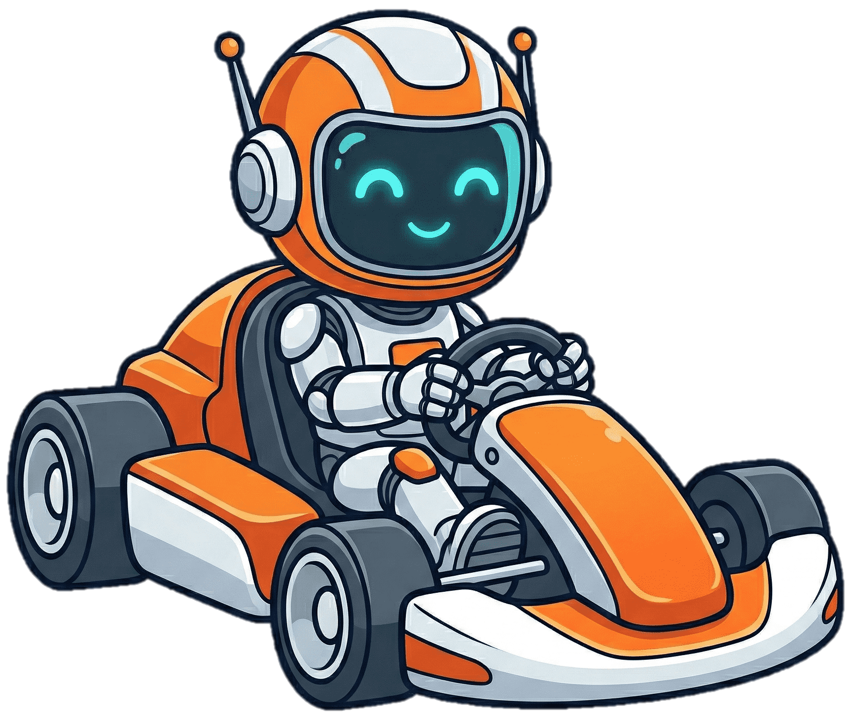}} \name{}: Unifying Understanding,
Perception, and Action Planning for Autonomous Driving}
\author[1,2]{Yongkang Li}
\author[2]{Lijun Zhou}
\author[1]{Sixu Yan}
\author[1]{Bencheng Liao}
\author[2,3]{Tianyi Yan}
\author[2]{Kaixin Xiong}
\author[2]{Long~Chen}
\author[2]{Hongwei Xie}
\author[2]{Bing Wang}
\author[2]{Guang Chen} 
\author[2]{Hangjun Ye}
\author[1]{Wenyu~Liu}
\author[2,\dagger]{Haiyang~Sun}
\author[1,\textrm{\Letter}]{Xinggang Wang}

\affiliation[1]{Huazhong University of Science and Technology}
\affiliation[2]{Xiaomi EV}
\affiliation[3]{SKL-IOTSC, University of Macau}

\contribution[\dagger]{Project Leader}
\contribution[\textrm{\Letter}]{Corresponding author}

\abstract{
    Vision-Language-Action (VLA) models have recently emerged in autonomous driving, with the promise of leveraging rich world knowledge to improve the cognitive capabilities of driving systems. However, adapting such models for driving tasks currently faces a critical dilemma between spatial perception and semantic reasoning. Consequently, existing VLA systems are forced into suboptimal compromises: directly adopting 2D Vision-Language Models yields limited spatial perception, whereas enhancing them with 3D spatial representations often impairs the native reasoning capacity of VLMs. We argue that this dilemma largely stems from the coupled optimization of spatial perception and semantic reasoning within shared model parameters. To overcome this, we propose \textbf{\name{}}, a \textbf{Uni}fied \textbf{Driv}ing \textbf{V}ision-\textbf{L}anguage-\textbf{A}ction model based on Mixture-of-Transformers that addresses the perception–reasoning conflict via expert decoupling. Specifically, it comprises three experts for driving understanding, scene perception, and action planning, which are coordinated through masked joint attention. In addition, we combine a sparse perception paradigm with a three-stage progressive training strategy to improve spatial perception while maintaining semantic reasoning capability. Extensive experiments show that \name{} achieves state-of-the-art performance in open-loop evaluation on nuScenes and closed-loop evaluation on Bench2Drive. Moreover, it demonstrates strong performance across a broad range of perception, prediction, and understanding tasks, including 3D detection, online mapping, motion forecasting, and driving-oriented VQA, highlighting its broad applicability as a unified model for autonomous driving.
}

\date{April 3, 2026}
\project{https://xiaomi-research.github.io/unidrivevla}
\code{https://github.com/xiaomi-research/unidrivevla}
\begin{document}
\thispagestyle{firstheader}
\maketitle

\section{Introduction}

\begin{center}
\small\itshape
One expert to understand them all, one expert to perceive them all, one expert to act for all.\footnotemark\\[-4pt]
\rule{0.72\linewidth}{0.4pt}
\end{center}
\footnotetext{Inspired by Tolkien's line ``One Ring to rule them all, One Ring to find them, One Ring to bring them all.''}

\begin{figure}[t!]
    \centering
    \includegraphics[width=\linewidth]{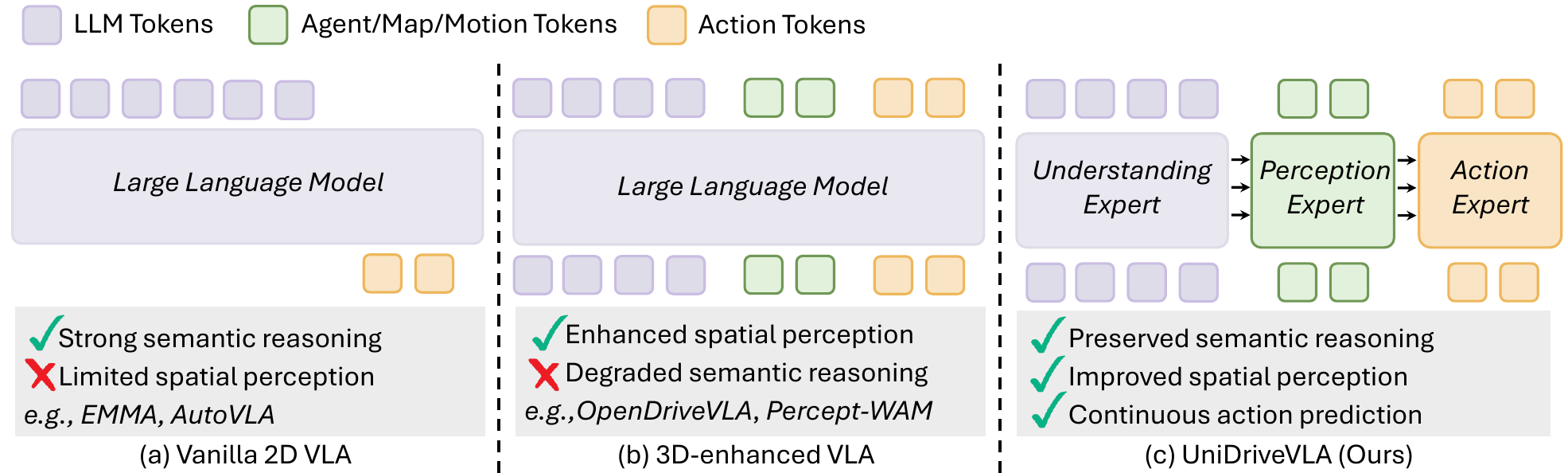}
    \caption{\textbf{Comparison of VLA paradigms for autonomous driving.} (a) Vanilla 2D VLA provides strong semantic reasoning but limited spatial perception. (b) 3D-enhanced VLA improves spatial perception but may degrade semantic reasoning. (c) \name{} decouples understanding, perception, and action with the Mixture-of-Transformers architecture, achieving both.}
    \label{fig:com}
    \vspace{-0.3cm}
\end{figure}

Autonomous driving has made significant progress in perception~\cite{li2024bevformer,liang2022bevfusion,liao2022maptr}, prediction~\cite{chai2019multipath,gu2023vip3d}, and planning~\cite{hu2023planning,jiang2023vad} in recent years.
Against this backdrop, Vision-Language-Action (VLA) models~\cite{wang2025alpamayo,li2025recogdrive,zhou2025autovla,hwang2024emma,fu2025orion} have emerged as a promising direction, leveraging rich world knowledge and reasoning capabilities to improve driving cognition and planning. However, their effectiveness largely depends on the underlying 2D Vision-Language Models (VLMs)~\cite{Qwen3-VL,zhu2025internvl3,islam2025gpt}. Because these base models are predominantly pre-trained on internet-scale image-text data, they are not designed for explicit spatial perception, which limits the spatial perception of VLA systems in driving scenarios. To improve spatial perception, recent works~\cite{dang2026sparseoccvla,liu2025occvla,zhou2025opendrivevla,fu2025orion,wang2025omnidrive,wang2026vggdrive} have explored enhancing VLA systems with 3D spatial representations, either by aligning structured 2D/3D features with language representations or by directly incorporating spatial tokens into a shared decoder. Yet this direction introduces a fundamental dilemma, as illustrated in Fig.~\ref{fig:com}: directly adopting 2D VLMs preserves native semantic reasoning but yields limited spatial perception, whereas enhancing them with 3D spatial representations improves spatial perception at the expense of native semantic reasoning.

At its core, this dilemma arises not merely from introducing 3D spatial representations, but from jointly optimizing spatial perception and semantic reasoning within shared model parameters. Specifically, although 3D spatial representations can improve spatial perception, jointly optimizing them with semantic reasoning in a shared parameter space introduces representation interference that undermines the native reasoning capacity of VLMs. However, most existing methods integrate spatial and semantic information within unified architectures with shared parameters. Among them, some recent approaches~\cite{wang2025omnidrive,zhou2025opendrivevla} attempt to mitigate this interference by explicitly aligning structured 2D/3D features with language representations, thereby introducing spatial cues while preserving semantic reasoning. Nevertheless, because such alignment data remains far smaller than the internet-scale corpora used for VLM pre-training, these methods can only partially mitigate the conflict and cannot fully eliminate it. Other works~\cite{han2025percept,liu2025occvla,dang2026sparseoccvla} instead introduce spatial cues by directly incorporating spatial tokens into a shared decoder, where spatial and semantic tokens are jointly modeled. As we later show in \figref{fig:cosine_com}, this observation is further supported by empirical analysis of representation similarity and downstream performance. Consequently, although these methods differ in how spatial information is introduced, they remain unable to fully reconcile spatial perception with semantic reasoning, highlighting the need to decouple these objectives during optimization.

To address this conflict, we propose \textbf{\name{}}, a unified Vision-Language-Action framework that unifies driving understanding, scene perception, and action planning within a single model. The core challenge is that simply decoupling these objectives is insufficient: the model must reduce optimization interference among heterogeneous semantic, spatial, and action tokens, while still preserving the unified decision process and the native semantic reasoning capability of the pre-trained VLM. To this end, \name{} adopts a Mixture-of-Transformers (MoT) architecture with three specialized experts for understanding, perception, and action, where optimization is decoupled across expert-specific pathways and masked joint attention maintains controlled cross-expert communication for coherent driving behavior. Moreover, instead of introducing dense 3D representations that may disrupt the native reasoning ability of VLMs, we adopt a sparse perception paradigm to extract critical spatial priors directly from 2D VLM tokens, thereby improving spatial perception while preserving semantic reasoning. Finally, to make this decoupled design trainable in practice, we introduce a three-stage progressive training strategy that progressively equips the model with perception and planning capabilities without sacrificing its foundational semantic understanding.

We evaluate \textbf{\name{}} on the widely used nuScenes and Bench2Drive benchmarks. Experimental results show that it achieves state-of-the-art performance in open-loop evaluation on nuScenes and closed-loop evaluation on Bench2Drive. Beyond driving planning, \name{} also performs well across a broad range of perception, prediction, and understanding tasks, including 3D detection, online mapping, motion forecasting, and driving-oriented visual question answering. These results highlight the advantage of our decoupled design: it improves spatial perception and driving performance while preserving the semantic reasoning ability of VLMs, demonstrating broad applicability as a unified model for autonomous driving.

\begin{itemize}[leftmargin=2.0em]

\item We propose \textbf{\name{}}, a pioneering unified Vision-Language-Action model based on a Mixture-of-Transformers architecture, which mitigates the conflict between spatial perception and semantic reasoning by decoupling their optimization with dedicated experts.

\item We introduce a sparse perception paradigm that extracts spatial priors directly from 2D VLM features, together with a three-stage joint training strategy that improves perception and planning while preserving the native reasoning ability of VLMs.

\item Extensive experiments on nuScenes and Bench2Drive show that \textbf{\name{}} achieves state-of-the-art performance in open-loop evaluation on nuScenes and closed-loop evaluation on Bench2Drive, while also demonstrating broad applicability across diverse perception, prediction, and understanding tasks.

\end{itemize}

\section{Related Work}

\vspace{-0.2cm}
\subsection{Vision-Language-Action Models for Autonomous Driving.}

Recent studies~\citep{hwang2024emma,fu2025orion,zeng2025futuresightdrive,zhou2025opendrivevla,yang2025drivemoe,wang2025alpamayo,li2025recogdrive,wang2025omnidrive} have explored integrating Vision-Language Models (VLMs) into autonomous driving, leveraging their rich world knowledge and reasoning capabilities to better address long-tail scenarios. Dual-system approaches, such as DriveVLM~\cite{tian2024drivevlm} and Senna~\cite{jiang2024senna}, adopt a fast-slow paradigm that combines VLMs with end-to-end (E2E) driving models~\cite{jiang2023vad,hu2023planning,sun2025sparsedrive}, where the VLM predicts low-frequency trajectories or provides high-level instructions to guide the E2E model. In contrast to these dual-system frameworks, recent methods have shifted toward unified single-system architectures~\cite{hwang2024emma,fu2025orion,zhou2025opendrivevla,zhou2025autovla,li2025recogdrive,zhang2024wisead,chi2025impromptu,xing2025openemma,li2025drivevla,renz2025simlingo}, enabling a single model to perform trajectory planning directly. To satisfy the high-frequency inference requirements of autonomous driving, several studies couple VLMs with action decoders~\cite{fu2025orion,jiang2025diffvla,renz2025simlingo,yang2025drivemoe,li2025drivevla,gao2025diffvla++,ang2026representational,wang2026hist,li2025recogdrive,li2026sgdrive,liu2026driveworld,schmidt2026lad,yan2025m} or employ Diffusion Large Language Models (DLLMs)~\cite{zeng2025diffusionvl,zhu2025llada,xu2025wam,dang2026drivefine,ma2025dvlm} to accelerate reasoning and support continuous action prediction. Recent works~\cite{zhou2025autovla,li2025recogdrive,wang2025alpamayo,fu2025minddrive,zhang2025openread,yan2025ad,xu2025wam,rawal2026nord,luo2025adathinkdrive,shang2026dynvla,luo2026last,luo2026unleashing} have also explored reinforcement learning (RL) for fine-tuning VLA models to further improve driving performance and policy alignment.

However, autonomous driving requires robust scene perception to recognize surrounding objects and road structures. Most existing VLA approaches~\cite{hwang2024emma,zhou2025autovla,wang2025alpamayo} primarily rely on fine-tuning pre-trained open-source models, whose spatial perception remains insufficient for such spatially demanding driving tasks. To improve spatial perception in VLA models, some approaches incorporate BEV encoders~\cite{zhou2025opendrivevla} or 3D Q-Formers~\cite{wang2025omnidrive,fu2025orion,bai20243d} to extract 3D features that are subsequently fed into LLMs. Alternatively, other methods~\cite{liu2025occvla,dang2026sparseoccvla,han2025percept} directly inject agent and occupancy tokens into the LLM decoder and jointly optimize the model for both linguistic and spatial perception objectives. Although these strategies enrich the model with 3D spatial information, they may also introduce interference between spatial perception and language reasoning in shared parameters, thereby weakening the effective use of pre-trained world knowledge. In contrast, our work focuses on reconciling spatial perception and semantic reasoning within a unified driving VLA framework.
\vspace{-0.2cm}
\subsection{Mixture-of-Transformers.}

The Mixture-of-Transformers (MoT)~\citep{liang2024mixture} paradigm was originally introduced to unify multiple modalities by integrating modality-decoupled experts within a Mixture-of-Experts (MoE)~\citep{jacobs1991adaptive} architecture together with decoupled attention mechanisms, thereby improving both modeling capacity and computational efficiency. Subsequent works~\citep{deng2025emerging,tang2025exploring,qin2025uni,luo2025next,shi2024lmfusion,marouani2026revisiting} extended MoT to multimodal understanding and generation tasks, where joint attention enables the incorporation of knowledge from Large Language Models (LLMs) into the generation process while preserving their comprehension capabilities, thereby improving generation performance. Concurrent with the emergence of Mixture-of-Transformers as a general architectural paradigm, $\pi_0$~\citep{black2024pi_0} independently adopted a conceptually similar mixture-based transformer design for robotics, integrating vision-language understanding with flow-matching action experts to unify discrete semantic reasoning and continuous action generation. Motivated by MoT's ability to mitigate modality interference through decoupled parameter spaces and computational sparsity, subsequent works following $\pi_0$ have adopted similar architectures in Vision-Language-Action (VLA) models for robotics~\cite{physicalintelligence2025pi05,cheang2025gr,team2025gigabrain,team2026gigabrain,luo2026being,huang2025motvla,cai2026xiaomi,li2026causal,yuan2025depthvla,bi2025motus,cai2026internvla} and autonomous driving~\cite{yang2025drivemoe,li2025drivevla,bartoccioni2025vavim,wang2025alpamayo,huang2026automot,zuo2026vega}.
However, in the autonomous driving domain, existing approaches~\cite{yang2025drivemoe,li2025drivevla,wang2025alpamayo} largely follow the $\pi_0$ design, yet their spatial perception remains limited, which can impair driving performance. In contrast, our work extends the MoT paradigm to autonomous driving by combining expert decoupling with sparse spatial perception, enabling a more effective unification of perception, reasoning, and planning.

\vspace{-0.2cm}
\subsection{Sparse Perception for Autonomous Driving.}
Traditional autonomous driving perception heavily relies on dense 3D representations~\cite{philion2020lift} or BEV grid constructions~\cite{li2023bevdepth,li2024bevformer,liang2022bevfusion}, which introduce computational redundancy. To circumvent this limitation, sparse perception models have emerged as an efficient alternative. These approaches adopt a query-driven paradigm~\cite{carion2020end}, using a set of sparse 3D queries to aggregate multi-view image features. Pioneering works such as DETR3D~\cite{wang2022detr3d} and PETR~\cite{liu2022petr} first introduced sparse object queries for 3D detection, bypassing dense view transformation. Subsequent advancements, including Sparse4D~\cite{lin2022sparse4d} and SparseBEV~\cite{liu2023sparsebev}, further bolstered spatial perception by integrating temporal modeling and adaptive spatio-temporal sampling mechanisms. More recently, this sparse paradigm has been extended to end-to-end autonomous driving systems. Methods such as SparseDrive~\cite{sun2025sparsedrive} and SparseAD~\cite{zhu2025sparsead} represent dynamic driving scenes using unified sparse queries and integrate detection, tracking, and planning within a single framework. Nevertheless, existing methods have not yet fully integrated sparse spatial perception with language-driven reasoning in VLA frameworks.

\vspace{-0.3cm}

\begin{figure}[t!]
    \centering
    \includegraphics[width=0.64\linewidth]{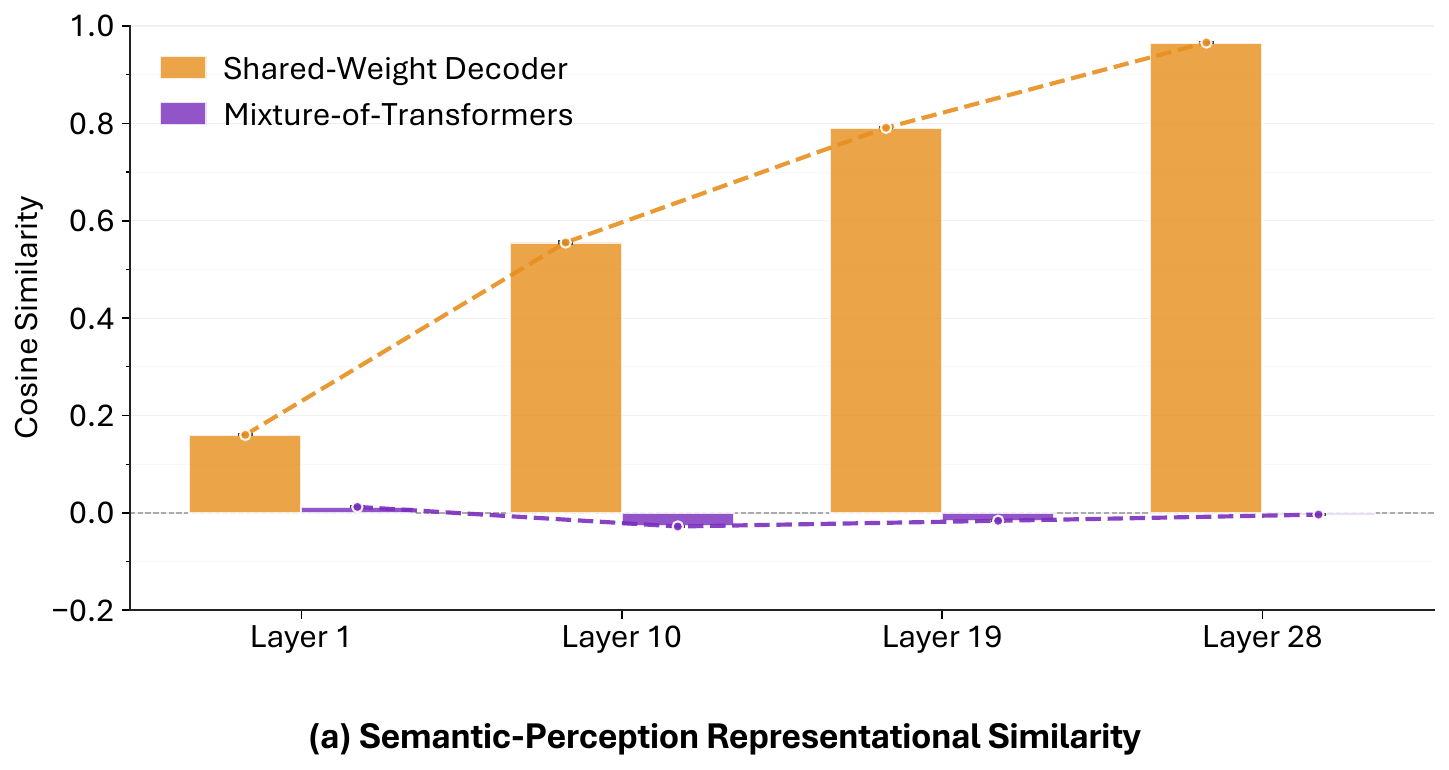}\hfill
    \includegraphics[width=0.35\linewidth]{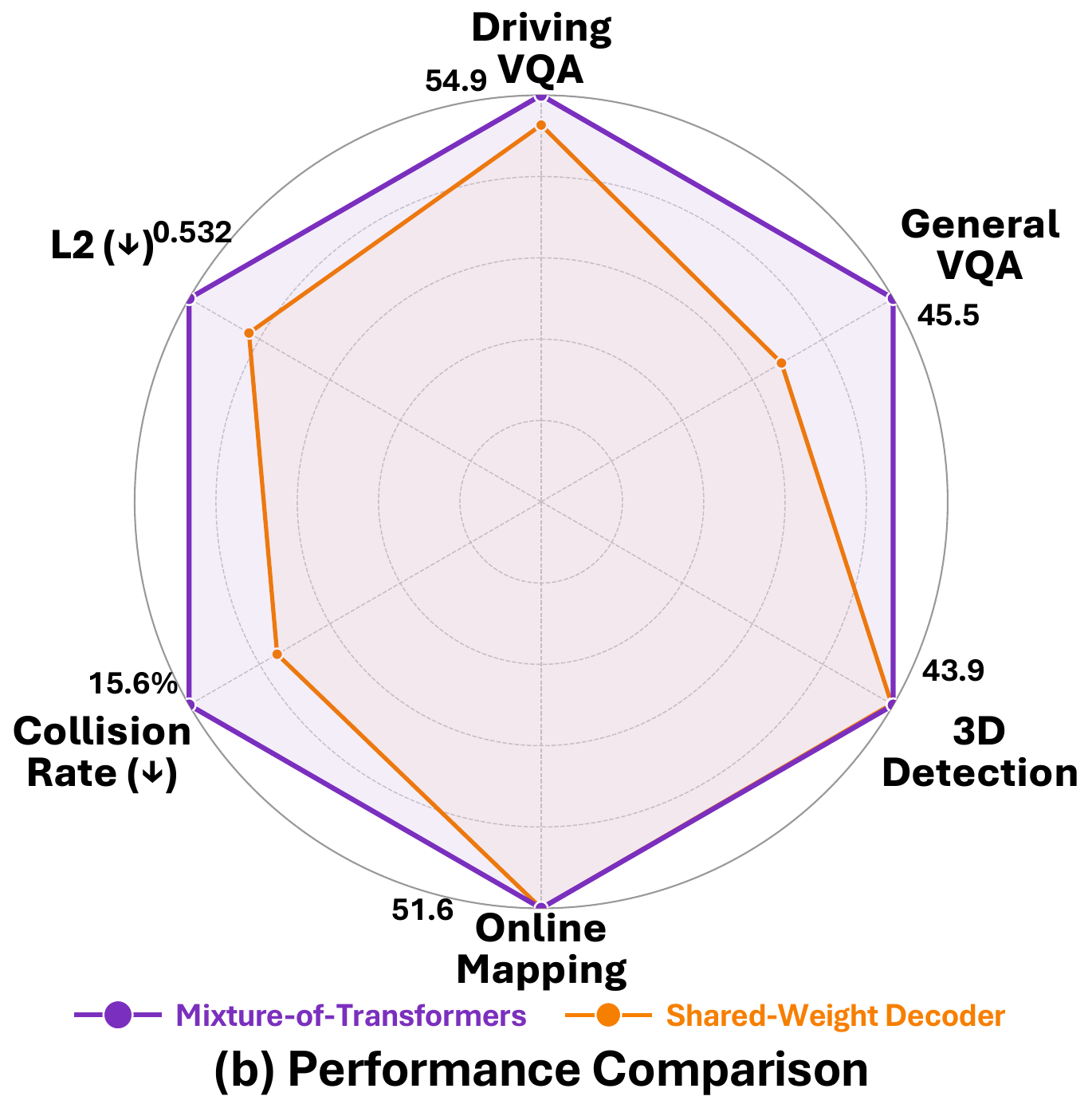}
    \caption{\textbf{Analysis of representation interference and model performance.} (a) Cosine similarity between LLM tokens and perception tokens across layers. In the shared-weight decoder, the similarity progressively increases toward 1, indicating feature collapse into nearly identical representations, whereas MoT maintains low similarity and preserves task decoupling. (b) Performance comparison. By mitigating optimization conflicts, \name{} consistently outperforms the shared-weight baseline across perception, reasoning, and planning metrics.}
    \label{fig:cosine_com}
    \vspace{-0.4cm}
\end{figure}
\section{Methodology}

\begin{figure}[t!]
    \centering
    \includegraphics[width=\linewidth]{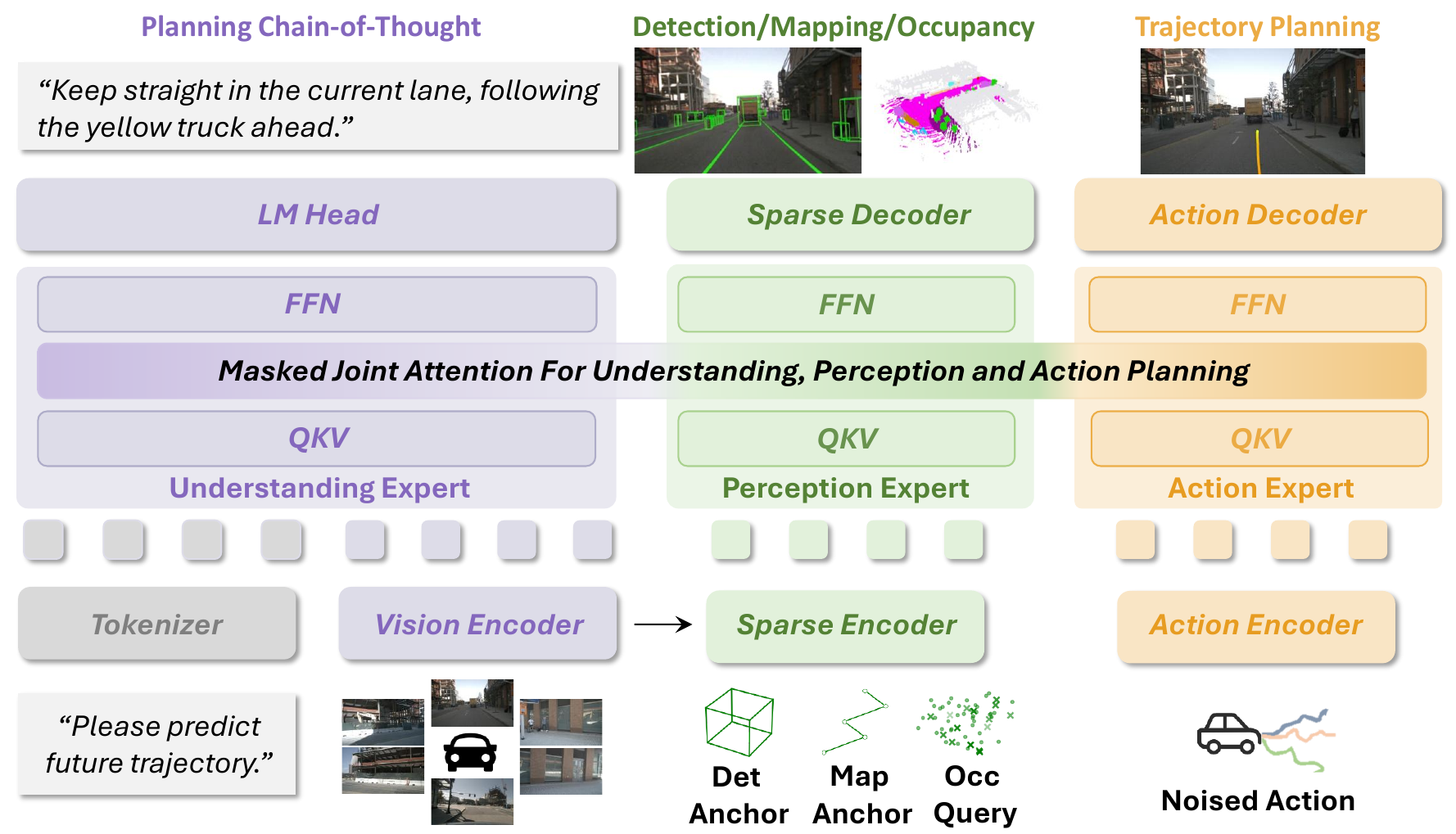}
    \caption{\textbf{Architecture overview of \name{}.} \name{} adopts a Mixture-of-Transformers architecture with three specialized experts for driving understanding, scene perception, and action planning. By decoupling heterogeneous tokens into dedicated experts and coordinating them through masked joint attention, the model mitigates optimization conflicts and unifies understanding, perception, and planning within a single framework.}
    \label{fig:model_arch}
    \vspace{-0.3cm}
\end{figure}

In this section, we present the methodology of \textbf{UniDriveVLA}, with an overview shown in Fig.~\ref{fig:model_arch}. We begin by formulating the perception--reasoning conflict in monolithic VLA models, where jointly optimizing spatial perception and semantic reasoning within shared parameters leads to representation interference. We then introduce the \textbf{UniDriveVLA} architecture based on a Mixture-of-Transformers (MoT) paradigm, which decouples the model into specialized experts for understanding, perception, and planning. Next, we describe a sparse perception mechanism that extracts spatial priors from 2D VLM features and supports downstream planning within the decoupled framework. Finally, we present a three-stage progressive training strategy that enables stable optimization of perception and planning without sacrificing the model's semantic capabilities.

\subsection{Problem Definition}

The autonomous driving task aims to predict a safe future trajectory $T_{\mathrm{traj}}$ from multi-view camera observations $I_{\mathrm{cam}} \in \mathbb{R}^{K\times V\times H\times W\times3}$, historical trajectory $I_{\mathrm{hist}} \in \mathbb{R}^{T_{\mathrm{hist}}\times2}$, and navigation command $L_{\mathrm{nav}}$.
\begin{equation}
    \label{eq:mono_vla}
    T_{\mathrm{traj}}=\Phi\bigl(I_{\mathrm{cam}},\,I_{\mathrm{hist}},\,L_{\mathrm{nav}}\bigr),
\end{equation}
where $T_{\mathrm{traj}}=\{(x_t,y_t)\}_{t=1}^{T}$ denotes the predicted future trajectory.

To improve spatial perception, recent 3D-enhanced VLA models further introduce spatial representations $T_{\mathrm{sp}}$, such as structured 2D/3D features or spatial tokens, into the same shared-weight decoder:
\begin{equation}
    \label{eq:mono_vla_spatial}
    T_{\mathrm{traj}}=\Phi\bigl(I_{\mathrm{cam}},\,I_{\mathrm{hist}},\,L_{\mathrm{nav}},\,T_{\mathrm{sp}}\bigr).
\end{equation}
However, although these spatial representations can enhance spatial perception, jointly modeling them with semantic reasoning within shared parameters introduces representation interference between heterogeneous features. As shown in \figref{fig:cosine_com}, this interference causes perception and semantic features to become increasingly entangled with depth, which weakens the native reasoning capacity of VLMs and ultimately degrades downstream driving performance.

\subsection{\name{} Architecture}

To address the perception--reasoning conflict in monolithic VLA models, \textbf{\name{}} adopts a Mixture-of-Transformers (MoT)~\cite{liang2024mixture} architecture that decouples the model into three specialized experts for \emph{driving understanding}, \emph{scene perception}, and \emph{action planning}. Instead of forcing heterogeneous tokens to share a single parameter space, our architecture assigns them to expert-specific pathways while retaining controlled cross-expert interaction through masked joint attention.

Given multi-view images $I_{\mathrm{cam}}$, historical trajectories $I_{\mathrm{hist}}$, and navigation command $L_{\mathrm{nav}}$, \name{} constructs three groups of tokens. First, a vision-language backbone produces understanding tokens $T_{\mathrm{und}}$ from visual observations and driving instructions. Second, a sparse perception module extracts perception tokens $T_{\mathrm{per}}$ from visual features to encode spatial priors. Third, an action encoder produces action tokens $T_{\mathrm{act}}$ for planning-oriented trajectory modeling, where the action inputs are constructed by standard flow matching interpolation between Gaussian noise and the target velocity sequence.

Within each MoT layer, the three token groups are first processed by expert-specific projections. For each expert $g \in \{\mathrm{und}, \mathrm{per}, \mathrm{act}\}$, the query, key, and value representations are computed as
\begin{equation}
    \mathbf{Q}^{g} = T_{g}\mathbf{W}_{Q}^{g}, \;
    \mathbf{K}^{g} = T_{g}\mathbf{W}_{K}^{g}, \;
    \mathbf{V}^{g} = T_{g}\mathbf{W}_{V}^{g}.
\end{equation}
This expert-specific parameterization decouples understanding, perception, and action into separate parameter subspaces before cross-expert interaction.

To coordinate information flow across experts, we introduce Masked Joint Attention. After expert-specific projection, the representations from all experts are concatenated in the order of understanding, perception, and action:
\begin{equation}
    \mathbf{Q} = [\mathbf{Q}^{\mathrm{und}};\, \mathbf{Q}^{\mathrm{per}};\, \mathbf{Q}^{\mathrm{act}}], \;
    \mathbf{K} = [\mathbf{K}^{\mathrm{und}};\, \mathbf{K}^{\mathrm{per}};\, \mathbf{K}^{\mathrm{act}}], \;
    \mathbf{V} = [\mathbf{V}^{\mathrm{und}};\, \mathbf{V}^{\mathrm{per}};\, \mathbf{V}^{\mathrm{act}}],
\end{equation}
and attention is computed globally with a mask matrix $\mathbf{M}$ that controls the visibility pattern:
\begin{equation}
    \mathbf{Z} = \mathrm{Softmax}\!\left(\frac{\mathbf{Q}\mathbf{K}^{\top}}{\sqrt{d_k}} + \mathbf{M}\right)\mathbf{V}.
\end{equation}
The resulting attention features are then split back into three expert groups,
\begin{equation}
    \mathbf{Z} = [\mathbf{Z}^{\mathrm{und}};\, \mathbf{Z}^{\mathrm{per}};\, \mathbf{Z}^{\mathrm{act}}],
\end{equation}
where $\mathbf{Z}^{\mathrm{und}}$, $\mathbf{Z}^{\mathrm{per}}$, and $\mathbf{Z}^{\mathrm{act}}$ correspond to the outputs associated with the understanding, perception, and action experts, respectively.

\begin{wrapfigure}[16]{r}{0.30\textwidth}
    \vspace{-6pt} 
    \centering
    \includegraphics[width=\linewidth]{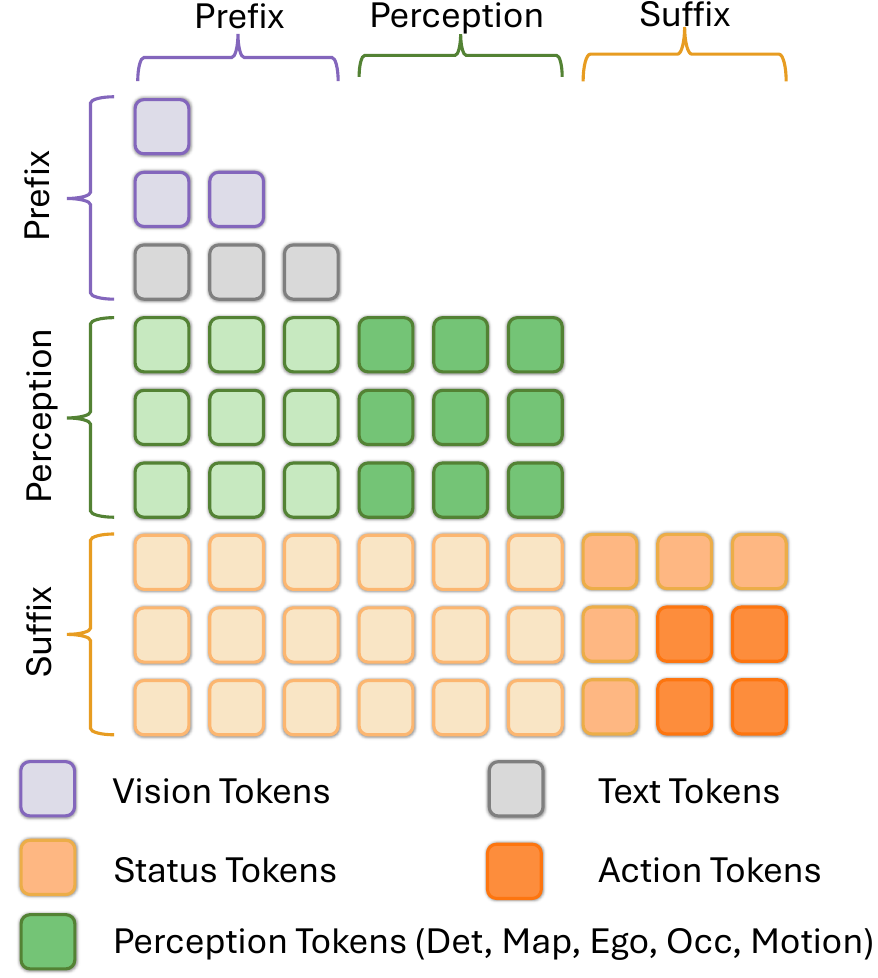}
    \vspace{-10pt}
    \caption{\textbf{Illustration of Masked Joint Attention.}}
    \label{fig:attention_mask}
    \vspace{-12pt}
\end{wrapfigure}
Here, understanding tokens follow causal masking and do not attend to subsequent perception or action tokens, preserving the semantic reasoning capability of the pre-trained vision-language model. Perception tokens attend to preceding understanding tokens to acquire semantic context, while action tokens aggregate both semantic and spatial information for planning. After attention, each expert group is further updated by expert-specific output projection, normalization, and feed-forward transformation:
\begin{equation}
    \mathbf{H}^{g} = T_{g} + \mathrm{LN}_{\mathrm{attn}}^{g}\!\left(\mathbf{Z}^{g}\mathbf{W}_{O}^{g}\right), \;
    \mathbf{O}^{g} = \mathbf{H}^{g} + \mathrm{LN}_{\mathrm{ffn}}^{g}\!\left(\mathrm{FFN}^{g}(\mathbf{H}^{g})\right),
\end{equation}

where $g \in \{\mathrm{und}, \mathrm{per}, \mathrm{act}\}$. The resulting expert features are then jointly optimized with a unified objective over understanding, perception, and action modeling within a single framework. Specifically, the understanding branch supports autoregressive language modeling, the perception branch is supervised by structured perception tasks, and the action branch is trained with flow-matching-based trajectory generation:

\begin{equation}
    \mathcal{L}_{\mathrm{total}} = \lambda_1 \mathcal{L}_{\mathrm{ar}} + \lambda_2 \mathcal{L}_{\mathrm{per}} + \lambda_3 \mathcal{L}_{\mathrm{act}}.
\end{equation}
In this way, \name{} unifies understanding, perception, and planning within a single framework while avoiding the optimization conflict of monolithic shared-parameter VLA models.

\subsection{Sparse Spatial Perception}

Unlike recent approaches~\cite{wang2025omnidrive,zhou2025opendrivevla} that inject dense bird's-eye-view representations into Vision-Language-Action models, \name{} constructs sparse spatial perception directly from multi-scale 2D visual features. Specifically, projected visual features serve as the geometric evidence for a unified query-based perception module that jointly models detection, online mapping, ego-status estimation, and motion forecasting, rather than assigning each task to an isolated prediction head. Task-specific sparse queries are initialized from instance banks obtained by dataset-level K-Means clustering and are then updated through temporal interaction, intra-task reasoning, inter-task communication, deformable feature aggregation, and task-wise refinement. In this way, the perception branch captures temporal dynamics, task-specific structure, and cross-task dependencies within a unified sparse decoding process. In parallel, occupancy is modeled as an auxiliary latent branch, so that the perception expert covers heterogeneous yet mutually supportive outputs, including 3D detection, map prediction, ego-status, motion, and occupancy.

To further enrich sparse perception with high-level semantics while preserving the native reasoning behavior of the pre-trained vision-language model, we project the first-pass perception outputs into the hidden space of the perception expert and let them interact with the understanding and action branches through masked joint attention. Concretely, detection, map, ego, motion, and occupancy tokens are lifted into the VLM hidden space, where understanding tokens remain causally masked, while perception and action tokens selectively aggregate preceding semantic and spatial context. As a result, perception tokens are semantically enhanced by the understanding branch and are further aggregated by the action branch to guide planning. The resulting features are then projected back to the sparse perception space and further refined by a subsequent perception decoder, yielding semantically enriched perception outputs that are better aligned with downstream planning. In this way, sparse perception in \name{} is not merely a one-shot geometric extractor, but a semantically enhanced perception module that collaborates with the understanding and action branches to guide planning, while also providing interpretable perception outputs.

\subsection{Three-Stage Progressive Training}

To preserve semantic reasoning while progressively acquiring spatial perception and planning capability, we devise a three-stage progressive training strategy for UniDriveVLA, which stabilizes optimization and mitigates catastrophic forgetting.

The first stage aims to anchor the model's semantic reasoning capability through large-scale multimodal pre-training. We follow established paradigms~\cite{li2025recogdrive} and curate a mixture of driving-specific visual question-answering datasets and general-domain multimodal data. By filtering low-quality driving dialogues and maintaining a dominant sampling ratio for high-quality general data, we preserve the foundational semantic capabilities of the vision-language model.

The second stage aims to incorporate perception and planning supervision into the model through controlled joint optimization. We perform joint optimization over autoregressive language modeling, spatial perception tasks including 3D detection, online mapping, and occupancy prediction, together with flow-matching-based trajectory generation. To preserve semantic reasoning capability during this controlled joint optimization, we apply Low-Rank Adaptation~\cite{hu2022lora} to the vision-language model. Furthermore, we reduce the base learning rate of the VLM parameters by half to suppress aggressive updates that may disrupt semantic reasoning.

The final stage aims to specialize the Perception Expert and Action Expert while preserving the semantic pathway of the vision-language model. We freeze the vision-language model and then fine-tune the Perception Expert and Action Expert while introducing an additional motion prediction objective. This objective is introduced to provide dynamic priors for the action expert and support motion-aware planning.
\section{Experiments} 
\subsection{Experimental Setup}  
\paragraph{Implementation Details.}
Our framework is built upon Qwen3-VL~\cite{Qwen3-VL}, a vision-language model consisting of a SigLIP-2~\cite{tschannen2025siglip} vision encoder, an MLP-based vision-language merger, and a Qwen3 LM~\cite{bai2025qwen3}. All input driving frames are resized to $960 \times 544$ to align with the $32\times$ spatial downsampling stride of the Qwen3-VL vision encoder. For the first stage of driving pre-training, following ReCogDrive~\cite{li2025recogdrive}, we fully fine-tune the base VLM for 3 epochs with a learning rate of $4\times10^{-5}$ on a mixture of driving-specific data and general vision-language data. The training mixture is constructed with a driving-to-general data ratio of $3{:}7$, where the general-domain portion mainly comes from FineVision~\cite{wiedmann2025finevision}. In the second stage, we jointly train the model for 30 epochs using AdamW~\cite{loshchilov2017decoupled}. The base learning rate is set to $2\times10^{-4}$, while the VLM backbone uses a $0.5\times$ learning-rate multiplier, yielding an effective learning rate of $1\times10^{-4}$. We additionally apply LoRA~\cite{hu2022lora} to the language model and use EMA during training. In the final stage, we freeze the vision-language model and fine-tune the Perception and Action Experts for 15 epochs, while enabling the motion forecasting objective. The base learning rate is set to $1\times10^{-4}$, and EMA is maintained during training.

\paragraph{Dataset.}
We primarily evaluate our framework on two widely adopted autonomous driving benchmarks: nuScenes~\cite{caesar2020nuscenes} and Bench2Drive~\cite{jia2024bench2drive}, which are used for open-loop and closed-loop evaluation, respectively. The nuScenes dataset consists of 1,000 driving sequences collected in Boston and Singapore, and we use it to evaluate both perception tasks and open-loop planning. Bench2Drive is a large-scale fully annotated benchmark built on the CARLA simulator. It provides 6-view camera inputs at a resolution of $900 \times 1600$ and is used for closed-loop driving evaluation. The expert demonstrations in Bench2Drive are officially generated using the Think2Drive~\cite{jia2023think} paradigm. In addition, we evaluate driving-oriented understanding on DriveBench~\cite{xie2025vlms}, and assess broader multimodal capability on several general VQA benchmarks, including MMStar~\cite{chen2024we}, MMMU~\cite{yue2024mmmu}, RealWorldQA, AI2D, MME~\cite{fu2023mme}, VLMsAreBlind~\cite{rahmanzadehgervi2024vision}, and ChartQA~\cite{masry2022chartqa}.

\begin{table*}[t]
    \centering
    \scriptsize
    \caption{
    Open-loop and closed-loop planning results on Bench2Drive. 
    Avg. L2 is computed over predictions within a 2-second horizon at 2 Hz. 
    $\dagger$~Models trained on PDM-Lite, which provides higher-quality expert demonstrations than Think2Drive. Bold numbers indicate the best results among methods trained without PDM-Lite.
    }\label{tab:planning_results}
    \setlength{\tabcolsep}{4pt} 
    \begin{tabular}{l c c c c c}
    \toprule
    \multirow{2}{*}{Method} & \multicolumn{1}{c}{Open-loop Metric} & \multicolumn{4}{c}{Closed-loop Metric} \\
    \cmidrule(lr){2-2} \cmidrule(lr){3-6}
    & Avg. L2 $\downarrow$ 
    & Driving Score $\uparrow$ 
    & Success Rate (\%) $\uparrow$ 
    & Efficiency $\uparrow$ 
    & Comfortness $\uparrow$ \\
    \midrule
    \pdm{AutoVLA\pdmnote~\pub{NeurIPS25}~\cite{zhou2025autovla}}  
    & \pdm{-} 
    & \pdm{78.84} 
    & \pdm{57.73} 
    & \pdm{146.93} 
    & \pdm{39.33} \\
    
    \pdm{SimLingo\pdmnote~\pub{CVPR25}~\cite{renz2025simlingo}}  
    & \pdm{-} 
    & \pdm{85.94} 
    & \pdm{66.82} 
    & \pdm{244.18} 
    & \pdm{25.49} \\
    
    \pdm{R2SE\pdmnote~\pub{T-PAMI26}~\cite{liu2026reinforced}}  
    & - & \pdm{86.28} & \pdm{69.54} & \pdm{243.89} & \pdm{23.26} \\
    \midrule
    AD-MLP~\pub{arXiv23}~\cite{zhai2023rethinking}                  & 3.64 & 18.05 & 0.00  & 48.45  & 22.63 \\
    VAD~\pub{ICCV23}~\cite{jiang2023vad}                     & 0.91 & 42.35 & 15.00 & 157.94 & 46.01 \\
    SparseDrive~\pub{ICRA25}~\cite{sun2025sparsedrive}            & 0.87 & 44.54 & 16.71 & 170.21 & 48.63 \\
    GenAD~\pub{ECCV24}~\cite{zheng2024genad}             & -   & 44.81 & 15.90 & -     & - \\
    UniAD~\pub{CVPR23}~\cite{hu2023planning}              & 0.73 & 45.81 & 16.36 & 129.21 & 43.58 \\
    MomAD~\pub{CVPR25}~\cite{song2025don}            & 0.82 & 47.91 & 18.11 & 174.91 & \textbf{51.20} \\
    SeerDrive~\pub{NeurIPS25}~\cite{zhang2025future}            & 0.66 & 58.32 & 30.17 & - & - \\
    DriveDPO~\pub{NeurIPS25}~\cite{shang2025drivedpo}            & - & 62.02 & 30.62 & 166.80 & 26.79 \\
    ThinkTwice~\pub{CVPR22}~\cite{jia2023think}     & 0.95 & 62.44 & 31.23 & 69.33 & 16.22 \\
    DriveTransformer~\pub{ICLR25}~\cite{jia2025drivetransformer}  & 0.62 & 63.46 & 35.01 & 100.64 & 20.78 \\
    DriveAdapter~\pub{ICCV23}~\cite{jia2023driveadapter}  & 1.01 & 64.22 & 33.08 & 70.22 & 16.01  \\
    RAP~\pub{ICLR26}~\cite{feng2025rap}  & - & 66.42 & 37.27 & 165.47 & 23.63 \\
    ReCogDrive~\pub{ICLR26}~\cite{li2025recogdrive}  & - & 71.36 & 45.45 & 138.18 & 17.45 \\
    DriveMOE~\pub{CVPR26}~\cite{yang2025drivemoe}  & \textbf{0.38} & 74.22 & 48.64 & 175.96 & 15.31 \\
    Orion~\pub{ICCV25}~\cite{fu2025orion}  & 0.68 & 77.74 & \textbf{54.62} & 151.48 & 17.38  \\
    \midrule
    \textbf{UniDriveVLA} & 0.72 & \textbf{78.37} & 51.82 & \textbf{198.86} & 11.78 \\
    \bottomrule
    \end{tabular}
\end{table*}

\begin{table*}[t]
    \centering
    \scriptsize
    \caption{Multi-ability results of end-to-end autonomous driving methods on Bench2Drive.}\label{tab:multi_ability_results}
    \begin{tabular}{l c c c c c c}
    \toprule
    \multirow{2}{*}{Method} 
    & \multicolumn{6}{c}{Ability (\%) $\uparrow$} \\
    \cmidrule(lr){2-7}
    & Merging 
    & Overtaking 
    & Emergency Brake 
    & Give Way 
    & Traffic Sign 
    & Mean \\
    \midrule
    AD-MLP~\pub{arXiv23}~\cite{zhai2023rethinking}                 & 0.00 & 0.00  & 0.00  & 0.00  & 4.35  & 0.87 \\
    UniAD~\pub{CVPR23}~\cite{hu2023planning}             & 14.10 & 17.78 & 21.67 & 10.00 & 14.21 & 15.55 \\
    VAD~\pub{ICCV23}~\cite{jiang2023vad}                    & 8.11 & 24.44 & 18.64 & 20.00 & 19.15 & 18.07 \\
    ThinkTwice~\pub{CVPR22}~\cite{jia2023think}    & 13.72 & 22.93 & 52.99 & \textbf{50.00} & 47.78 & 37.48 \\
    DriveAdapter~\pub{ICCV23}~\cite{jia2023driveadapter}  & 14.55 & 22.61 & 54.04 & \textbf{50.00} & 50.45 & 38.33 \\
    DriveTransformer~\pub{ICLR25}~\cite{jia2025drivetransformer} & 17.57 & 35.00 & 48.36 & 40.00 & 52.10 & 38.60 \\
    ReCogDrive~\pub{ICLR26}~\cite{li2025recogdrive}  & 29.73 & 20.00 & 69.09 & 20.00 & 71.34 & 42.03  \\
    DriveMOE~\pub{CVPR26}~\cite{yang2025drivemoe} & 34.67 & 40.00 & 65.45 & 40.00 & 59.44 & 47.91 \\
    Orion~\pub{ICCV25}~\cite{fu2025orion}  & 25.00 & 71.11 & \textbf{78.33} & 33.00 & \textbf{69.15} & \textbf{54.72} \\
    \midrule
    \textbf{UniDriveVLA} 
    & \textbf{38.75}
    &  \textbf{80.00}
    &  50.00
    &  30.00
    &  58.95
    &  51.53 \\
    \bottomrule
    \end{tabular}
\end{table*}

\subsection{Main Results and Ablation Study}

\paragraph{Experiments on the Bench2Drive Benchmark.}
We evaluate UniDriveVLA on Bench2Drive from both planning performance and fine-grained driving ability. As shown in \tabref{tab:planning_results}, UniDriveVLA achieves the best Driving Score of 78.37 and the best Efficiency of 198.86 among methods trained without PDM-Lite, while obtaining a competitive Success Rate of 51.82\%. In \tabref{tab:multi_ability_results}, UniDriveVLA demonstrates strong capability across diverse interactive scenarios, achieving the best performance on Merging with 38.75\% and Overtaking with 80.00\%, together with a competitive mean score of 51.53\%. These results indicate that UniDriveVLA not only improves closed-loop driving quality, but also handles complex interactive behaviors effectively.

\begin{table*}[t]
    \centering
    \caption{\textbf{End-to-end trajectory planning performance on nuScenes.} We evaluate planning quality using L2 displacement error and collision rate across two mainstream protocols: ST-P3~\cite{hu2022st} and UniAD~\cite{hu2023planning}. * denotes methods using ego-state information. Results for SparseDrive$^{\ddagger}$ are re-evaluated following the protocol established by GPT-Driver~\cite{mao2023gpt}.}
    \vspace{-0.5em}
    \resizebox{\textwidth}{!}{ 
    \begin{tabular}{lcccccccc|cccccccc|c}
    \toprule
     \multirow{4}{*}{\textbf{Method}} & \multicolumn{8}{c}{\textbf{ST-P3 metrics}} & \multicolumn{8}{c}{\textbf{UniAD metrics}} & \multirow{4}{*}{\textbf{LLM}}   \\
    \cmidrule(lr){2-9} \cmidrule(lr){10-17}
    & \multicolumn{4}{c}{\textbf{L2 (m)} $\downarrow$} & \multicolumn{4}{c}{\textbf{Collision (\%)} $\downarrow$} & \multicolumn{4}{c}{\textbf{L2 (m)} $\downarrow$} & \multicolumn{4}{c}{\textbf{Collision (\%)} $\downarrow$} &   \\
    \cmidrule(lr){2-5} \cmidrule(lr){6-9} \cmidrule(lr){10-13} \cmidrule(lr){14-17}
    & 1s & 2s & 3s & Avg. & 1s & 2s & 3s & Avg. & 1s & 2s & 3s & Avg. & 1s & 2s & 3s & Avg. &  \\
    \midrule
    \multicolumn{18}{c}{\textbf{Methods with Ego Status}} \\
    \midrule
    ST-P3*~\cite{hu2022st} & 1.33 & 2.11 & 2.90 & 2.11 & 0.23 & 0.62 & 1.27 & 0.71 & - & - & - & - & - & - & - & - & -  \\ 
    VAD*~\cite{jiang2023vad} & 0.17 & 0.34 & 0.60 & 0.37 & 0.04 & 0.27 & 0.67 & 0.33 & - & - & - & - & - & - & - & - & -  \\
    UniAD*~\cite{hu2023planning} & - & - & - & - & - & - & - & - & 0.20 & 0.42 & \textbf{0.75} & 0.46 & 0.02 & 0.25 & 0.84 & 0.37 & -  \\
    RDA-Driver*~\cite{huang2024making} & 0.17 & 0.37 & 0.69 & 0.40 & 0.01 & \textbf{0.05} & 0.26 & 0.10 & 0.23 & 0.73 & 1.54 & 0.80 & \textbf{0.00} & 0.13 & 0.83 & 0.32 & LLaVA-7B  \\
    BEV-Planner*~\cite{li2024ego} & 0.16 & 0.32 & 0.57 & 0.35 & \textbf{0.00} & 0.29 & 0.73 & 0.34 & - & - & - & - & - & - & - & - & -  \\
    HPP*~\cite{liu2025hybrid} & 0.26 & 0.37 & 0.59 & 0.40 & 0.02 & \textbf{0.05} & \textbf{0.11} & \textbf{0.06} & 0.30 & 0.61 & 1.15 & 0.72 & 0.03 & 0.07 & \textbf{0.35} & \textbf{0.15} & -  \\
    OmniDrive*~\cite{wang2025omnidrive} & \textbf{0.14}  & 0.29 & 0.55 & 0.33 & \textbf{0.00} & 0.13 & 0.78 & 0.30  & - & - & - & - & - & - & - & - & LLaVA-7B \\
    Orion*~\cite{fu2025orion} & 0.17  & 0.31 & 0.55 & 0.34 & 0.05 & 0.25 & 0.80 & 0.37  & - & - & - & - & - & - & - & - & LLaVA-7B \\
    FSDrive*$^{\dag}$~\cite{zeng2025futuresightdrive} & \textbf{0.14} & \textbf{0.25} & \textbf{0.46} & \textbf{0.28} & 0.03 & 0.06 & 0.21 & 0.10 & \textbf{0.18} & \textbf{0.39} & 0.77 & \textbf{0.45} & \textbf{0.00} & \textbf{0.06} & 0.42 & 0.16  & Qwen2-VL-3B  \\ 
    AutoVLA*~\cite{zhou2025autovla} & 0.25  & 0.46 & 0.73 & 0.48 & 0.07 & 0.07 & 0.26 & 0.13 & 0.33 & 0.81 & 1.45 & 0.86 & 0.08 & 0.11 & 0.85 & 0.35 & Qwen2.5-VL-3B \\
    OpenDriveVLA*~\cite{zhou2025opendrivevla} & \textbf{0.14}  & 0.30 & 0.55 & 0.33 & 0.02 & 0.07 & 0.22 & 0.10 & 0.19 & 0.58 & 1.24 & 0.67 & 0.02 & 0.18 & 0.70 & 0.30 & Qwen2.5-VL-3B \\
    \midrule
    \textbf{UniDriveVLA-Base*} & 0.23 & 0.40 & 0.65 & 0.43 & 0.04 & 0.09 & 0.18 & 0.10 & 0.30 & 0.69 & 1.32 & 0.77 & 0.03 & 0.13 & 0.52 & 0.23 & Qwen3-VL-2B \\
    \textbf{UniDriveVLA-Large*} & 0.24 & 0.40 & 0.63 & 0.42 & 0.03 & 0.09 & 0.16 & 0.10 & 0.30 & 0.66 & 1.25 & 0.74 & 0.02 & 0.18 & 0.40 & 0.20 & Qwen3-VL-8B \\
    \midrule
    \multicolumn{18}{c}{\textbf{Methods without Ego Status}} \\
    \midrule
    VAD~\cite{jiang2023vad} & 0.41 & 0.70 & 1.05 & 0.72 & \textbf{0.03} & 0.19 & 0.43 & 0.21 & - & - & - & - & - & - & - & - & -  \\
    UniAD~\cite{hu2023planning} & 0.45 & 0.70 & 1.04 & 0.73 & 0.62 & 0.58 & 0.63 & 0.61 & 0.59 & 1.01 & 1.48 & 1.03 & 0.16 & 0.51 & 1.64 & 0.77 & -  \\
    ELM~\cite{zhou2024embodied}& - & - & - & - & - & - & - & -& \textbf{0.34}  & 1.23 & 2.57 & 1.38 & 0.12 & 0.50 & 2.36 & 0.99 & BLIP2-2.7B\\
    OccWorld~\cite{zheng2024occworld} & 0.39  & 0.73 & 1.18 & 0.77 & 0.11 & 0.19 & 0.67 & 0.32 & 0.52  & 1.27 & 2.41 & 1.40 & 0.12 & 0.40 & 2.08 & 0.87 & GPT3-like\\
    BEV-Planner~\cite{li2024ego} & 0.30 & 0.52 & 0.83 & 0.55 & 0.10 & 0.37 & 1.30 & 0.59 & - & - & - & - & - & - & - & - & -  \\
    HPP~\cite{liu2025hybrid} & 0.41 & 0.61 & 0.86 & 0.63 & \textbf{0.03} & 0.08 & 0.24 & 0.12 & 0.48 & 0.91 & 1.54 & 0.97 & 0.03 & 0.17 & 0.68 & 0.29 & -  \\
    SparseDrive$^{\ddagger}$~\cite{sun2025sparsedrive} & 0.28 & 0.53 & 0.84 & 0.55 & \textbf{0.03} & \textbf{0.05} & \textbf{0.15} & \textbf{0.08} & 0.38 & 0.92 & 1.66 & 0.99 & \textbf{0.02} & \textbf{0.08} & \textbf{0.53} & \textbf{0.21} & -  \\
    OmniDrive~\cite{wang2025omnidrive} & 0.40  & 0.80 & 1.32 & 0.84 & 0.04 & 0.46 & 2.32 & 0.94 & - & - & - & - & - & - & - & - & LLaVA-7B \\
    FSDrive~\cite{zeng2025futuresightdrive} & 0.28 & 0.52 & 0.80 & 0.53 & 0.06 & 0.13 & 0.32 & 0.17 & 0.40 & 0.89 & 1.60 & 0.96 & 0.07 & 0.12 & 1.02 & 0.40  & Qwen2-VL-3B  \\ 
    \midrule
    \textbf{UniDriveVLA-Base} & 0.28 & 0.51 & 0.82 & 0.54 & 0.08 & 0.13 & 0.31 & 0.17 & 0.37 & 0.89 & 1.62 & 0.96 & 0.08 & 0.27 & 0.88 & 0.41 & Qwen3-VL-2B \\
    \textbf{UniDriveVLA-Large} & \textbf{0.27} & \textbf{0.49} & \textbf{0.77} & \textbf{0.51} & \textbf{0.03} & 0.10 & 0.21 & 0.11 & 0.36 & \textbf{0.83} & \textbf{1.50} & \textbf{0.90} & \textbf{0.02} & 0.23 & 0.55 & 0.27 & Qwen3-VL-8B \\
    \bottomrule
    \end{tabular}}
    \label{tab:nuscenes_planning}
\end{table*}

\begin{table}[t]
    \centering
    \scriptsize
    \setlength{\tabcolsep}{4pt}
    \renewcommand{\arraystretch}{1.05}
    \caption{Comparison of perception, mapping, and motion prediction performance on the nuScenes validation dataset.}
    \begin{tabular}{lcccccccc}
    \toprule
    \multirow{2}{*}{Method}
    & \multicolumn{2}{c}{Detection}
    & \multicolumn{4}{c}{Map}
    & \multicolumn{2}{c}{Motion} \\
    \cmidrule(lr){2-3}
    \cmidrule(lr){4-7}
    \cmidrule(lr){8-9}
    & mAP$\uparrow$
    & NDS$\uparrow$
    & AP$_{ped}\uparrow$
    & AP$_{divider}\uparrow$
    & AP$_{boundary}\uparrow$
    & mAP$\uparrow$
    & minADE$(m)\downarrow$
    & minFDE$(m)\downarrow$ \\
    \midrule

    UniAD~\pub{CVPR23}~\cite{hu2023planning}
    & 0.380 & 0.359
    & -- & -- & -- & --
    & 0.710 & 1.020 \\

    VAD~\pub{ICCV23}~\cite{jiang2023vad}
    & 0.276 & 0.397
    & 0.406 & 0.515 & 0.506 & 0.476
    & -- & -- \\

    SparseDrive~\pub{ICRA25}~\cite{sun2025sparsedrive}
    & 0.418 & 0.525
    & 0.499 & \textbf{0.570} & \textbf{0.584} & 0.551
    & \textbf{0.600} & \textbf{0.960} \\

    EgoFSD~\pub{ICRA26}~\cite{su2024difsd}
    & 0.410 & 0.528
    & \textbf{0.549} & 0.557 & 0.573 & 0.560
    & -- & -- \\

    HiP-AD~\pub{ICCV25}~\cite{tang2025hip}
    & \textbf{0.424} & \textbf{0.535}
    & -- & -- & -- & \textbf{0.571}
    & 0.610 & -- \\

    \midrule
    \textbf{UniDriveVLA-Base}
    & 0.397 & 0.434
    & 0.462 & 0.556 & 0.543 & 0.520
    & 1.396 & 2.289 \\
    \textbf{UniDriveVLA-Large}
    & 0.407 & 0.460
    & 0.491 & 0.556 & 0.557 & 0.535
    & 1.264 & 2.121 \\
    \bottomrule
    \end{tabular}
    \label{tab:perception}
\end{table}

\begin{table*}[t]
\centering

\begin{minipage}[t]{0.50\textwidth}
\centering
\captionof{table}{Ablation on nuScenes planning. L2 and CR denote average L2 error and collision rate, respectively.}
\label{tab:abl}
\vspace{0pt}
\resizebox{\linewidth}{!}{
\begin{tabular}{cccccc|cc}
\toprule
\multicolumn{6}{c|}{Components} & \multicolumn{2}{c}{Metrics} \\
\cmidrule(lr){1-6} \cmidrule(lr){7-8}
Baseline & Ego & Det & Map & Occ & Motion & L2 $\downarrow$ & CR (\%) $\downarrow$ \\
\midrule
\checkmark &  &  &  &  &  & 0.75 & 0.27 \\
\checkmark & \checkmark &  &  &  &  & 0.61 & 0.21 \\
\checkmark & \checkmark & \checkmark &  &  &  & 0.58 & \textbf{0.10} \\
\checkmark & \checkmark & \checkmark & \checkmark &  &  & 0.58 & 0.14 \\
\checkmark & \checkmark & \checkmark & \checkmark & \checkmark &  & \textbf{0.53} & 0.14 \\
\checkmark & \checkmark & \checkmark & \checkmark & \checkmark & \checkmark & 0.54 & 0.17 \\
\bottomrule
\end{tabular}
}
\end{minipage}
\hfill
\begin{minipage}[t]{0.46\textwidth}
\centering
\captionof{table}{DriveBench comparison. $\dagger$ ReCogDrive is pretrained only, without action training.}
\label{tab:drivebench_only}
\vspace{0pt}
\resizebox{\linewidth}{!}{
\begin{tabular}{lccccc}
\toprule
\textbf{Method} & \textbf{Percep.} & \textbf{Predict.} & \textbf{Plan.} & \textbf{Behav.} & \textbf{Avg.} \\
\midrule
LLaVA-1.5~\cite{liu2023visual}
& 23.22 & 22.02 & 29.15 & 13.60 & 22.00 \\

InternVL2~\cite{team2024internvl2}
& 32.36 & 45.52 & 53.27 & 54.58 & 46.43 \\

Qwen2-VL~\cite{wang2024qwen2}
& 30.13 & 49.35 & 61.30 & 51.26 & 48.01 \\

DriveLM~\cite{sima2024drivelm}
& 16.85 & 44.33 & 68.71 & 42.78 & 43.17 \\

Dolphins~\cite{ma2024dolphins}
& 9.59 & 32.66 & 52.91 & 8.81 & 25.99 \\

GPT-4o~\cite{islam2025gpt}
& 35.37 & \textbf{51.30} & \textbf{75.75} & 45.40 & 51.96 \\

ReCogDrive$^{\dagger}$~\cite{li2025recogdrive}
& \textbf{64.95} & 49.34 & 70.20 & 42.36 & \textbf{56.71} \\

\midrule
\textbf{UniDriveVLA}
& 36.78 & 43.13 & 66.98 & \textbf{60.97} & 51.97 \\
\bottomrule
\end{tabular}
}
\end{minipage}

\end{table*}

\begin{table*}[t]
    \centering
    \caption{Performance comparison between a shared-weight decoder and Mixture-of-Transformers (MoT) across understanding, perception, and planning tasks. General VQA is averaged over general multimodal benchmarks, DriveBench is used for driving-oriented understanding, and perception and planning metrics are reported on nuScenes.}
    \vspace{0pt}
    \resizebox{\linewidth}{!}{
    \begin{tabular}{l|cc|cc|cc}
    \toprule
    \multirow{2}{*}{\textbf{Architecture}} & \multicolumn{2}{c|}{\textbf{Understanding}} & \multicolumn{2}{c|}{\textbf{Perception}} & \multicolumn{2}{c}{\textbf{Planning}} \\
    \cmidrule(lr){2-3} \cmidrule(lr){4-5} \cmidrule(lr){6-7}
    & General VQA (\%) $\uparrow$ & DriveBench (\%) $\uparrow$ & Det (NDS) $\uparrow$ & Map (mAP) $\uparrow$ & L2 (m) $\downarrow$ & CR (\%) $\downarrow$ \\
    \midrule
    Shared-Weight Decoder & 31.1 & 50.8 & 0.437 & \textbf{0.516} & 0.641 & 0.175 \\
    Mixture-of-Transformers & \textbf{45.5} & \textbf{54.9} & \textbf{0.439} & \textbf{0.516} & \textbf{0.533} & \textbf{0.140} \\
    \bottomrule
    \end{tabular}
    }
    \label{tab:mot_vs_shared}
\end{table*}

\begin{table*}[!t]
    \centering
    \scriptsize
    \setlength{\tabcolsep}{4pt}
    \caption{Comparison of UniDriveVLA with general-purpose and driving-specific models on general multimodal understanding benchmarks.}
    \label{tab:general_vqa}
    \begin{tabular}{l c c c c c c c c}
    \toprule
    \textbf{Model} & \textbf{Params} & \textbf{MMStar}$\uparrow$ & \textbf{MMMU}$\uparrow$ & \textbf{RealWorldQA}$\uparrow$ & \textbf{AI2D}$\uparrow$ & \textbf{MME}$\uparrow$ & \textbf{VLMsAreBlind}$\uparrow$ & \textbf{ChartQA}$\uparrow$ \\
    \midrule
    
    Qwen2.5-VL~\cite{bai1others} & 7B & 63.7 & 51.1 & \textbf{69.0} & 82.5 & 2317 & 40.3 & 82.5 \\
    
    Qwen3-VL~\cite{bai2025qwen3} & 8B & 63.0 & 52.8 & \textbf{69.0} & 83.2 & 2364 & \textbf{61.9} & 83.8 \\
    
    InternVL3~\cite{zhu2025internvl3} & 8B & \textbf{69.1} & 54.8 & 67.8 & \textbf{83.9} & \textbf{2393} & 42.0 & 86.6 \\

    InternVL3.5~\cite{wang2025internvl3} & 8B & 66.3 & 57.3 & 63.0 & 82.3 & 2357 & 45.4 & \textbf{87.1} \\
            
    GPT-5 Nano~\cite{singh2025openai} & — & 41.3 & \textbf{57.6} & 60.7 & 65.7 & - & 40.2 & 48.6 \\
    \midrule
    \textbf{UniDriveVLA} & 8B & 43.3 & 47.3 & 49.9 & 76.3 & 1876 & 26.6 & 76.3 \\
    \bottomrule
    \end{tabular}
\end{table*}






\paragraph{Planning Results on the nuScenes Benchmark.}
\tabref{tab:nuscenes_planning} presents a comprehensive comparison of end-to-end trajectory planning performance on nuScenes under two evaluation protocols, namely ST-P3 and UniAD. Overall, UniDriveVLA achieves competitive results in both the with-ego and without-ego settings. In particular, under the more challenging setting without ego-state inputs, UniDriveVLA-Large attains the best trajectory accuracy, achieving the lowest average L2 error under both the ST-P3 and UniAD protocols. In the with-ego setting, UniDriveVLA also remains competitive against recent VLA methods such as AutoVLA and FSDrive, as well as strong end-to-end driving baselines including UniAD and VAD. These results suggest that integrating sparse perception priors into a vision-language-action framework can substantially improve open-loop trajectory prediction, especially when explicit ego-state information is unavailable.

\paragraph{Perception Results on the nuScenes Benchmark.}
Beyond planning, we evaluate the intermediate perception capabilities of UniDriveVLA on the nuScenes validation set. As shown in \tabref{tab:perception}, UniDriveVLA achieves competitive detection and mapping performance. In particular, UniDriveVLA-Large reaches a detection mAP of 0.407 and an NDS of 0.460, while achieving a map mAP of 0.535. Compared with earlier end-to-end driving methods such as VAD, our model shows clear improvements in both detection and mapping quality, indicating that the proposed sparse perception design can provide a solid spatial foundation for downstream planning. Although its motion prediction accuracy remains behind specialized baselines, these results show that UniDriveVLA maintains meaningful multi-task perception capability within a unified VLA framework.

\paragraph{Ablation on Planning Components.}
We progressively incorporate ego-state, detection, mapping, occupancy, and motion components into the planning framework, and report the results in \tabref{tab:abl}. Starting from the baseline planner, introducing ego-state yields a substantial improvement in both trajectory accuracy and collision rate. Adding detection further reduces the collision rate to 0.10, indicating that explicit object-level perception is particularly beneficial for safe planning. Incorporating occupancy brings the best L2 error of 0.53, suggesting that dense spatial context is especially helpful for trajectory prediction. In contrast, adding mapping and motion does not lead to further gains in the current setting. We attribute this mainly to the limited headroom of open-loop planning on nuScenes, where improvements are already marginal and harder to realize.

\paragraph{Driving Scene Understanding.}
We evaluate UniDriveVLA on DriveBench to assess driving-oriented scene understanding, and further report results on several general VQA benchmarks to examine whether the model preserves multimodal capability after driving adaptation. As shown in \tabref{tab:drivebench_only}, UniDriveVLA achieves strong performance on DriveBench, demonstrating driving-oriented understanding and behavior reasoning.

\paragraph{Effect of MoT Decoupling.}
\tabref{tab:mot_vs_shared} compares a shared-weight decoder with the proposed Mixture-of-Transformers. MoT consistently improves understanding, perception, and planning performance, with especially clear gains on trajectory prediction and collision reduction. These results support our claim that expert decoupling effectively mitigates the perception--reasoning conflict in unified driving VLA models.

\paragraph{General Visual Capabilities.}
A key challenge in driving-oriented adaptation is preserving the general multimodal capability of the foundation model. As shown in \tabref{tab:general_vqa}, although UniDriveVLA does not match strong general-purpose VLMs on these benchmarks, it retains meaningful performance across a diverse set of general visual understanding tasks, achieving 49.9 on RealWorldQA, 76.3 on AI2D, and 76.3 on ChartQA. These results suggest that our driving adaptation does not collapse the model's multimodal reasoning ability, and that the adopted training strategy preserves a level of visual capability beyond autonomous driving.

\section{Conclusion}

In this work, we present \textbf{UniDriveVLA}, a unified Vision-Language-Action framework for autonomous driving that unifies understanding, perception, and planning within a single model. We identify a fundamental perception--reasoning conflict in existing monolithic VLA models, where jointly optimizing spatial perception and semantic reasoning within shared parameters leads to representation interference. To address this issue, we introduce a Mixture-of-Transformers architecture with dedicated Understanding, Perception, and Action experts, enabling controlled cross-expert interaction while reducing optimization conflict. We further adopt a sparse perception mechanism that extracts spatial priors directly from multi-scale 2D visual features, together with a three-stage progressive training strategy that improves perception and planning while preserving semantic reasoning capability. Extensive experiments on perception, understanding, and planning benchmarks demonstrate the effectiveness of UniDriveVLA as a unified model for autonomous driving. Moreover, the proposed decoupled VLA design may be further extended to robotic manipulation scenarios that require both structured spatial perception and high-level semantic reasoning.

\section*{Acknowledgments}

We use publicly available datasets, models, and code resources, including DriveLM, LingoQA, nuScenes, Bench2Drive, and other public research resources. We confirm that all such resources are used solely for academic research purposes and not for any commercial activity, in accordance with their respective licenses and terms of use.

We also thank Tianheng Cheng, Haoyu Fu, Lunbin Zeng, Hongyuan Tao, and Haochen Liu for their valuable discussions and helpful exchanges.

{
    \small
    \bibliographystyle{ieeenat_fullname}
    \bibliography{paper}

@String(ICLR = {Int. Conf. Learn. Represent.})

@String(AAAI = {AAAI})

@String(ICLR  = {ICLR})

@article{liang2024mixture,
  title={Mixture-of-transformers: A sparse and scalable architecture for multi-modal foundation models},
  author={Liang, Weixin and Yu, Lili and Luo, Liang and Iyer, Srinivasan and Dong, Ning and Zhou, Chunting and Ghosh, Gargi and Lewis, Mike and Yih, Wen-tau and Zettlemoyer, Luke and others},
  journal={arXiv preprint arXiv:2411.04996},
  year={2024}
}

@article{jacobs1991adaptive,
  title={Adaptive mixtures of local experts},
  author={Jacobs, Robert A and Jordan, Michael I and Nowlan, Steven J and Hinton, Geoffrey E},
  journal={Neural computation},
  volume={3},
  number={1},
  pages={79--87},
  year={1991},
  publisher={MIT Press}
}

@article{deng2025emerging,
  title={Emerging properties in unified multimodal pretraining},
  author={Deng, Chaorui and Zhu, Deyao and Li, Kunchang and Gou, Chenhui and Li, Feng and Wang, Zeyu and Zhong, Shu and Yu, Weihao and Nie, Xiaonan and Song, Ziang and others},
  journal={arXiv preprint arXiv:2505.14683},
  year={2025}
}

@inproceedings{tang2025exploring,
  title={Exploring the deep fusion of large language models and diffusion transformers for text-to-image synthesis},
  author={Tang, Bingda and Zheng, Boyang and Paul, Sayak and Xie, Saining},
  booktitle={Proceedings of the Computer Vision and Pattern Recognition Conference},
  pages={28586--28595},
  year={2025}
}

@article{qin2025uni,
  title={Uni-cot: Towards unified chain-of-thought reasoning across text and vision},
  author={Qin, Luozheng and Gong, Jia and Sun, Yuqing and Li, Tianjiao and Yang, Mengping and Yang, Xiaomeng and Qu, Chao and Tan, Zhiyu and Li, Hao},
  journal={arXiv preprint arXiv:2508.05606},
  year={2025}
}

@article{luo2025next,
  title={Next-omni: Towards any-to-any omnimodal foundation models with discrete flow matching},
  author={Luo, Run and Xia, Xiaobo and Wang, Lu and Chen, Longze and Shan, Renke and Luo, Jing and Yang, Min and Chua, Tat-Seng},
  journal={arXiv preprint arXiv:2510.13721},
  year={2025}
}

@article{shi2024lmfusion,
  title={Lmfusion: Adapting pretrained language models for multimodal generation},
  author={Shi, Weijia and Han, Xiaochuang and Zhou, Chunting and Liang, Weixin and Lin, Xi Victoria and Zettlemoyer, Luke and Yu, Lili},
  journal={arXiv preprint arXiv:2412.15188},
  year={2024}
}

@article{black2024pi_0,
  title={$pi\_0 $: A Vision-Language-Action Flow Model for General Robot Control},
  author={Black, Kevin and Brown, Noah and Driess, Danny and Esmail, Adnan and Equi, Michael and Finn, Chelsea and Fusai, Niccolo and Groom, Lachy and Hausman, Karol and Ichter, Brian and others},
  journal={arXiv preprint arXiv:2410.24164},
  year={2024}
}

@article{physicalintelligence2025pi05,
  title   = {{$\pi_{0.5}$}: A Vision-Language-Action Model with Open-World Generalization},
  author  = {{Physical Intelligence} and Black, Kevin and Brown, Noah and Darpinian, James and Dhabalia, Karan and Driess, Danny and Esmail, Adnan and Equi, Michael and Finn, Chelsea and Fusai, Niccolo and others},
  journal = {arXiv preprint arXiv:2504.16054},
  year    = {2025}
}

@article{cheang2025gr,
  title={Gr-3 technical report},
  author={Cheang, Chilam and Chen, Sijin and Cui, Zhongren and Hu, Yingdong and Huang, Liqun and Kong, Tao and Li, Hang and Li, Yifeng and Liu, Yuxiao and Ma, Xiao and others},
  journal={arXiv preprint arXiv:2507.15493},
  year={2025}
}

@article{team2025gigabrain,
  title={Gigabrain-0: A world model-powered vision-language-action model},
  author={Team, GigaBrain and Ye, Angen and Wang, Boyuan and Ni, Chaojun and Huang, Guan and Zhao, Guosheng and Li, Haoyun and Li, Jie and Zhu, Jiagang and Feng, Lv and others},
  journal={arXiv preprint arXiv:2510.19430},
  year={2025}
}

@article{team2026gigabrain,
  title={GigaBrain-0.5 M*: a VLA That Learns From World Model-Based Reinforcement Learning},
  author={Team, GigaBrain and Wang, Boyuan and Ni, Chaojun and Huang, Guan and Zhao, Guosheng and Li, Hao and Li, Jie and Lv, Jindi and Liu, Jingyu and Feng, Lv and others},
  journal={arXiv preprint arXiv:2602.12099},
  year={2026}
}

@article{luo2026being,
  title={Being-H0. 5: Scaling Human-Centric Robot Learning for Cross-Embodiment Generalization},
  author={Luo, Hao and Wang, Ye and Zhang, Wanpeng and Zheng, Sipeng and Xi, Ziheng and Xu, Chaoyi and Xu, Haiweng and Yuan, Haoqi and Zhang, Chi and Wang, Yiqing and others},
  journal={arXiv preprint arXiv:2601.12993},
  year={2026}
}

@article{huang2025motvla,
  title={MoTVLA: A Vision-Language-Action Model with Unified Fast-Slow Reasoning},
  author={Huang, Wenhui and Chen, Changhe and Qi, Han and Lv, Chen and Du, Yilun and Yang, Heng},
  journal={arXiv preprint arXiv:2510.18337},
  year={2025}
}

@article{cai2026xiaomi,
  title={Xiaomi-Robotics-0: An Open-Sourced Vision-Language-Action Model with Real-Time Execution},
  author={Cai, Rui and Guo, Jun and He, Xinze and Jin, Piaopiao and Li, Jie and Lin, Bingxuan and Liu, Futeng and Liu, Wei and Ma, Fei and Ma, Kun and others},
  journal={arXiv preprint arXiv:2602.12684},
  year={2026}
}

@article{li2026causal,
  title={Causal World Modeling for Robot Control},
  author={Li, Lin and Zhang, Qihang and Luo, Yiming and Yang, Shuai and Wang, Ruilin and Han, Fei and Yu, Mingrui and Gao, Zelin and Xue, Nan and Zhu, Xing and others},
  journal={arXiv preprint arXiv:2601.21998},
  year={2026}
}

@article{yang2025drivemoe,
  title={DriveMoE: Mixture-of-experts for vision-language-action model in end-to-end autonomous driving},
  author={Yang, Zhenjie and Chai, Yilin and Jia, Xiaosong and Li, Qifeng and Shao, Yuqian and Zhu, Xuekai and Su, Haisheng and Yan, Junchi},
  journal={arXiv preprint arXiv:2505.16278},
  year={2025}
}

@article{li2025drivevla,
  title={DriveVLA-W0: World models amplify data scaling law in autonomous driving},
  author={Li, Yingyan and Shang, Shuyao and Liu, Weisong and Zhan, Bing and Wang, Haochen and Wang, Yuqi and Chen, Yuntao and Wang, Xiaoman and An, Yasong and Tang, Chufeng and others},
  journal={arXiv preprint arXiv:2510.12796},
  year={2025}
}

@article{bartoccioni2025vavim,
  title={Vavim and vavam: Autonomous driving through video generative modeling},
  author={Bartoccioni, Florent and Ramzi, Elias and Besnier, Victor and Venkataramanan, Shashanka and Vu, Tuan-Hung and Xu, Yihong and Chambon, Loick and Gidaris, Spyros and Odabas, Serkan and Hurych, David and others},
  journal={arXiv preprint arXiv:2502.15672},
  year={2025}
}

@article{yuan2025depthvla,
  title={Depthvla: Enhancing vision-language-action models with depth-aware spatial reasoning},
  author={Yuan, Tianyuan and Liu, Yicheng and Lu, Chenhao and Chen, Zhuoguang and Jiang, Tao and Zhao, Hang},
  journal={arXiv preprint arXiv:2510.13375},
  year={2025}
}

@article{bi2025motus,
  title={Motus: A unified latent action world model},
  author={Bi, Hongzhe and Tan, Hengkai and Xie, Shenghao and Wang, Zeyuan and Huang, Shuhe and Liu, Haitian and Zhao, Ruowen and Feng, Yao and Xiang, Chendong and Rong, Yinze and others},
  journal={arXiv preprint arXiv:2512.13030},
  year={2025}
}

@article{cai2026internvla,
  title={InternVLA-A1: Unifying Understanding, Generation and Action for Robotic Manipulation},
  author={Cai, Junhao and Cai, Zetao and Cao, Jiafei and Chen, Yilun and He, Zeyu and Jiang, Lei and Li, Hang and Li, Hengjie and Li, Yang and Liu, Yufei and others},
  journal={arXiv preprint arXiv:2601.02456},
  year={2026}
}

@article{hwang2024emma,
  title={Emma: End-to-end multimodal model for autonomous driving},
  author={Hwang, Jyh-Jing and Xu, Runsheng and Lin, Hubert and Hung, Wei-Chih and Ji, Jingwei and Choi, Kristy and Huang, Di and He, Tong and Covington, Paul and Sapp, Benjamin and others},
  journal={arXiv preprint arXiv:2410.23262},
  year={2024}
}

@inproceedings{fu2025orion,
  title={Orion: A holistic end-to-end autonomous driving framework by vision-language instructed action generation},
  author={Fu, Haoyu and Zhang, Diankun and Zhao, Zongchuang and Cui, Jianfeng and Liang, Dingkang and Zhang, Chong and Zhang, Dingyuan and Xie, Hongwei and Wang, Bing and Bai, Xiang},
  booktitle={Proceedings of the IEEE/CVF International Conference on Computer Vision},
  pages={24823--24834},
  year={2025}
}

@article{zeng2025futuresightdrive,
  title={Futuresightdrive: Thinking visually with spatio-temporal cot for autonomous driving},
  author={Zeng, Shuang and Chang, Xinyuan and Xie, Mengwei and Liu, Xinran and Bai, Yifan and Pan, Zheng and Xu, Mu and Wei, Xing and Guo, Ning},
  journal={arXiv preprint arXiv:2505.17685},
  year={2025}
}

@article{zhou2025opendrivevla,
  title={Opendrivevla: Towards end-to-end autonomous driving with large vision language action model},
  author={Zhou, Xingcheng and Han, Xuyuan and Yang, Feng and Ma, Yunpu and Tresp, Volker and Knoll, Alois},
  journal={arXiv preprint arXiv:2503.23463},
  year={2025}
}

@inproceedings{caesar2020nuscenes,
  title={nuscenes: A multimodal dataset for autonomous driving},
  author={Caesar, Holger and Bankiti, Varun and Lang, Alex H and Vora, Sourabh and Liong, Venice Erin and Xu, Qiang and Krishnan, Anush and Pan, Yu and Baldan, Giancarlo and Beijbom, Oscar},
  booktitle={Proceedings of the IEEE/CVF conference on computer vision and pattern recognition},
  pages={11621--11631},
  year={2020}
}

@inproceedings{hu2022st,
  title={St-p3: End-to-end vision-based autonomous driving via spatial-temporal feature learning},
  author={Hu, Shengchao and Chen, Li and Wu, Penghao and Li, Hongyang and Yan, Junchi and Tao, Dacheng},
  booktitle={European Conference on Computer Vision},
  pages={533--549},
  year={2022},
  organization={Springer}
}

@inproceedings{hu2023planning,
  title={Planning-oriented autonomous driving},
  author={Hu, Yihan and Yang, Jiazhi and Chen, Li and Li, Keyu and Sima, Chonghao and Zhu, Xizhou and Chai, Siqi and Du, Senyao and Lin, Tianwei and Wang, Wenhai and others},
  booktitle={Proceedings of the IEEE/CVF conference on computer vision and pattern recognition},
  pages={17853--17862},
  year={2023}
}

@inproceedings{jiang2023vad,
  title={Vad: Vectorized scene representation for efficient autonomous driving},
  author={Jiang, Bo and Chen, Shaoyu and Xu, Qing and Liao, Bencheng and Chen, Jiajie and Zhou, Helong and Zhang, Qian and Liu, Wenyu and Huang, Chang and Wang, Xinggang},
  booktitle={Proceedings of the IEEE/CVF International Conference on Computer Vision},
  pages={8340--8350},
  year={2023}
}

@inproceedings{li2024ego,
  title={Is ego status all you need for open-loop end-to-end autonomous driving?},
  author={Li, Zhiqi and Yu, Zhiding and Lan, Shiyi and Li, Jiahan and Kautz, Jan and Lu, Tong and Alvarez, Jose M},
  booktitle={Proceedings of the IEEE/CVF Conference on Computer Vision and Pattern Recognition},
  pages={14864--14873},
  year={2024}
}

@inproceedings{huang2024making,
  title={Making large language models better planners with reasoning-decision alignment},
  author={Huang, Zhijian and Tang, Tao and Chen, Shaoxiang and Lin, Sihao and Jie, Zequn and Ma, Lin and Wang, Guangrun and Liang, Xiaodan},
  booktitle={European Conference on Computer Vision},
  pages={73--90},
  year={2024},
  organization={Springer}
}

@inproceedings{wang2025omnidrive,
  title={Omnidrive: A holistic vision-language dataset for autonomous driving with counterfactual reasoning},
  author={Wang, Shihao and Yu, Zhiding and Jiang, Xiaohui and Lan, Shiyi and Shi, Min and Chang, Nadine and Kautz, Jan and Li, Ying and Alvarez, Jose M},
  booktitle={Proceedings of the computer vision and pattern recognition conference},
  pages={22442--22452},
  year={2025}
}

@article{zhou2025autovla,
  title={Autovla: A vision-language-action model for end-to-end autonomous driving with adaptive reasoning and reinforcement fine-tuning},
  author={Zhou, Zewei and Cai, Tianhui and Zhao, Seth Z and Zhang, Yun and Huang, Zhiyu and Zhou, Bolei and Ma, Jiaqi},
  journal={arXiv preprint arXiv:2506.13757},
  year={2025}
}

@inproceedings{zhou2024embodied,
  title={Embodied understanding of driving scenarios},
  author={Zhou, Yunsong and Huang, Linyan and Bu, Qingwen and Zeng, Jia and Li, Tianyu and Qiu, Hang and Zhu, Hongzi and Guo, Minyi and Qiao, Yu and Li, Hongyang},
  booktitle={European Conference on Computer Vision},
  pages={129--148},
  year={2024},
  organization={Springer}
}

@inproceedings{zheng2024occworld,
  title={Occworld: Learning a 3d occupancy world model for autonomous driving},
  author={Zheng, Wenzhao and Chen, Weiliang and Huang, Yuanhui and Zhang, Borui and Duan, Yueqi and Lu, Jiwen},
  booktitle={European conference on computer vision},
  pages={55--72},
  year={2024},
  organization={Springer}
}

@article{liu2025hybrid,
  title={Hybrid-prediction integrated planning for autonomous driving},
  author={Liu, Haochen and Huang, Zhiyu and Huang, Wenhui and Yang, Haohan and Mo, Xiaoyu and Lv, Chen},
  journal={IEEE Transactions on Pattern Analysis and Machine Intelligence},
  volume={47},
  number={4},
  pages={2597--2614},
  year={2025},
  publisher={IEEE}
}

@inproceedings{sun2025sparsedrive,
  title={Sparsedrive: End-to-end autonomous driving via sparse scene representation},
  author={Sun, Wenchao and Lin, Xuewu and Shi, Yining and Zhang, Chuang and Wu, Haoran and Zheng, Sifa},
  booktitle={2025 IEEE International Conference on Robotics and Automation (ICRA)},
  pages={8795--8801},
  year={2025},
  organization={IEEE}
}

@article{zhai2023rethinking,
  title={Rethinking the open-loop evaluation of end-to-end autonomous driving in nuscenes},
  author={Zhai, Jiang-Tian and Feng, Ze and Du, Jinhao and Mao, Yongqiang and Liu, Jiang-Jiang and Tan, Zichang and Zhang, Yifu and Ye, Xiaoqing and Wang, Jingdong},
  journal={arXiv preprint arXiv:2305.10430},
  year={2023}
}

@inproceedings{renz2025simlingo,
  title={Simlingo: Vision-only closed-loop autonomous driving with language-action alignment},
  author={Renz, Katrin and Chen, Long and Arani, Elahe and Sinavski, Oleg},
  booktitle={Proceedings of the Computer Vision and Pattern Recognition Conference},
  pages={11993--12003},
  year={2025}
}

@inproceedings{zheng2024genad,
  title={Genad: Generative end-to-end autonomous driving},
  author={Zheng, Wenzhao and Song, Ruiqi and Guo, Xianda and Zhang, Chenming and Chen, Long},
  booktitle={European Conference on Computer Vision},
  pages={87--104},
  year={2024},
  organization={Springer}
}

@inproceedings{song2025don,
  title={Don't shake the wheel: Momentum-aware planning in end-to-end autonomous driving},
  author={Song, Ziying and Jia, Caiyan and Liu, Lin and Pan, Hongyu and Zhang, Yongchang and Wang, Junming and Zhang, Xingyu and Xu, Shaoqing and Yang, Lei and Luo, Yadan},
  booktitle={Proceedings of the IEEE/CVF Conference on Computer Vision and Pattern Recognition},
  pages={22432--22441},
  year={2025}
}

@article{zhang2025future,
  title={Future-aware end-to-end driving: Bidirectional modeling of trajectory planning and scene evolution},
  author={Zhang, Bozhou and Song, Nan and Li, Jingyu and Zhu, Xiatian and Deng, Jiankang and Zhang, Li},
  journal={arXiv preprint arXiv:2510.11092},
  year={2025}
}

@article{shang2025drivedpo,
  title={Drivedpo: Policy learning via safety dpo for end-to-end autonomous driving},
  author={Shang, Shuyao and Chen, Yuntao and Wang, Yuqi and Li, Yingyan and Zhang, Zhaoxiang},
  journal={arXiv preprint arXiv:2509.17940},
  year={2025}
}

@inproceedings{jia2023think,
  title={Think twice before driving: Towards scalable decoders for end-to-end autonomous driving},
  author={Jia, Xiaosong and Wu, Penghao and Chen, Li and Xie, Jiangwei and He, Conghui and Yan, Junchi and Li, Hongyang},
  booktitle={Proceedings of the IEEE/CVF Conference on Computer Vision and Pattern Recognition},
  pages={21983--21994},
  year={2023}
}

@article{jia2025drivetransformer,
  title={Drivetransformer: Unified transformer for scalable end-to-end autonomous driving},
  author={Jia, Xiaosong and You, Junqi and Zhang, Zhiyuan and Yan, Junchi},
  journal={arXiv preprint arXiv:2503.07656},
  year={2025}
}

@article{feng2025rap,
  title={Rap: 3d rasterization augmented end-to-end planning},
  author={Feng, Lan and Gao, Yang and Zablocki, Eloi and Li, Quanyi and Li, Wuyang and Liu, Sichao and Cord, Matthieu and Alahi, Alexandre},
  journal={arXiv preprint arXiv:2510.04333},
  year={2025}
}

@article{li2025recogdrive,
  title={Recogdrive: A reinforced cognitive framework for end-to-end autonomous driving},
  author={Li, Yongkang and Xiong, Kaixin and Guo, Xiangyu and Li, Fang and Yan, Sixu and Xu, Gangwei and Zhou, Lijun and Chen, Long and Sun, Haiyang and Wang, Bing and others},
  journal={arXiv preprint arXiv:2506.08052},
  year={2025}
}

@inproceedings{tang2025hip,
  title={Hip-ad: Hierarchical and multi-granularity planning with deformable attention for autonomous driving in a single decoder},
  author={Tang, Yingqi and Xu, Zhuoran and Meng, Zhaotie and Cheng, Erkang},
  booktitle={Proceedings of the IEEE/CVF International Conference on Computer Vision},
  pages={25605--25615},
  year={2025}
}

@article{liu2026reinforced,
  title={Reinforced refinement with self-aware expansion for end-to-end autonomous driving},
  author={Liu, Haochen and Li, Tianyu and Yang, Haohan and Chen, Li and Wang, Caojun and Guo, Ke and Tian, Haochen and Li, Hongchen and Li, Hongyang and Lv, Chen},
  journal={IEEE Transactions on Pattern Analysis and Machine Intelligence},
  year={2026},
  publisher={IEEE}
}

@inproceedings{jia2023driveadapter,
  title={Driveadapter: Breaking the coupling barrier of perception and planning in end-to-end autonomous driving},
  author={Jia, Xiaosong and Gao, Yulu and Chen, Li and Yan, Junchi and Liu, Patrick Langechuan and Li, Hongyang},
  booktitle={Proceedings of the IEEE/CVF International Conference on Computer Vision},
  pages={7953--7963},
  year={2023}
}

@article{su2024difsd,
  title={DiFSD: Ego-centric fully sparse paradigm with uncertainty denoising and iterative refinement for efficient end-to-end self-driving},
  author={Su, Haisheng and Wu, Wei and Yan, Junchi},
  journal={arXiv preprint arXiv:2409.09777},
  year={2024}
}

@article{zhu2025internvl3,
  title={Internvl3: Exploring advanced training and test-time recipes for open-source multimodal models},
  author={Zhu, Jinguo and Wang, Weiyun and Chen, Zhe and Liu, Zhaoyang and Ye, Shenglong and Gu, Lixin and Tian, Hao and Duan, Yuchen and Su, Weijie and Shao, Jie and others},
  journal={arXiv preprint arXiv:2504.10479},
  year={2025}
}

@article{wang2025internvl3,
  title={Internvl3. 5: Advancing open-source multimodal models in versatility, reasoning, and efficiency},
  author={Wang, Weiyun and Gao, Zhangwei and Gu, Lixin and Pu, Hengjun and Cui, Long and Wei, Xingguang and Liu, Zhaoyang and Jing, Linglin and Ye, Shenglong and Shao, Jie and others},
  journal={arXiv preprint arXiv:2508.18265},
  year={2025}
}

@article{bai2025qwen3,
  title={Qwen3-vl technical report},
  author={Bai, Shuai and Cai, Yuxuan and Chen, Ruizhe and Chen, Keqin and Chen, Xionghui and Cheng, Zesen and Deng, Lianghao and Ding, Wei and Gao, Chang and Ge, Chunjiang and others},
  journal={arXiv preprint arXiv:2511.21631},
  year={2025}
}

@article{bai1others,
  title={others. 2025. Qwen2. 5-vl technical report},
  author={Bai, Shuai and Chen, Keqin and Liu, Xuejing and Wang, Jialin and Ge, Wenbin and Song, Sibo and Dang, Kai and Wang, Peng and Wang, Shijie and Tang, Jun and others},
  journal={arXiv preprint arXiv:2502.13923},
  volume={4},
  number={5},
  year={1}
}

@article{singh2025openai,
  title={Openai gpt-5 system card},
  author={Singh, Aaditya and Fry, Adam and Perelman, Adam and Tart, Adam and Ganesh, Adi and El-Kishky, Ahmed and McLaughlin, Aidan and Low, Aiden and Ostrow, AJ and Ananthram, Akhila and others},
  journal={arXiv preprint arXiv:2601.03267},
  year={2025}
}

@inproceedings{islam2025gpt,
  title={Gpt-4o: The cutting-edge advancement in multimodal llm},
  author={Islam, Raisa and Moushi, Owana Marzia},
  booktitle={Intelligent Computing-Proceedings of the Computing Conference},
  pages={47--60},
  year={2025},
  organization={Springer}
}

@misc{team2024internvl2,
  title={Internvl2: Better than the best—expanding performance boundaries of open-source multimodal models with the progressive scaling strategy},
  author={Team, OpenGVLab},
  year={2024},
  publisher={Accessed}
}

@article{liu2023visual,
  title={Visual instruction tuning},
  author={Liu, Haotian and Li, Chunyuan and Wu, Qingyang and Lee, Yong Jae},
  journal={Advances in neural information processing systems},
  volume={36},
  pages={34892--34916},
  year={2023}
}

@article{wang2024qwen2,
  title={Qwen2-vl: Enhancing vision-language model's perception of the world at any resolution},
  author={Wang, Peng and Bai, Shuai and Tan, Sinan and Wang, Shijie and Fan, Zhihao and Bai, Jinze and Chen, Keqin and Liu, Xuejing and Wang, Jialin and Ge, Wenbin and others},
  journal={arXiv preprint arXiv:2409.12191},
  year={2024}
}

@inproceedings{sima2024drivelm,
  title={Drivelm: Driving with graph visual question answering},
  author={Sima, Chonghao and Renz, Katrin and Chitta, Kashyap and Chen, Li and Zhang, Hanxue and Xie, Chengen and Bei{\ss}wenger, Jens and Luo, Ping and Geiger, Andreas and Li, Hongyang},
  booktitle={European conference on computer vision},
  pages={256--274},
  year={2024},
  organization={Springer}
}

@inproceedings{ma2024dolphins,
  title={Dolphins: Multimodal language model for driving},
  author={Ma, Yingzi and Cao, Yulong and Sun, Jiachen and Pavone, Marco and Xiao, Chaowei},
  booktitle={European Conference on Computer Vision},
  pages={403--420},
  year={2024},
  organization={Springer}
}

@article{wang2025alpamayo,
  title={Alpamayo-r1: Bridging reasoning and action prediction for generalizable autonomous driving in the long tail},
  author={Wang, Yan and Luo, Wenjie and Bai, Junjie and Cao, Yulong and Che, Tong and Chen, Ke and Chen, Yuxiao and Diamond, Jenna and Ding, Yifan and Ding, Wenhao and others},
  journal={arXiv preprint arXiv:2511.00088},
  year={2025}
}

@article{zhang2025openread,
  title={OpenREAD: Reinforced Open-Ended Reasoning for End-to-End Autonomous Driving with LLM-as-Critic},
  author={Zhang, Songyan and Huang, Wenhui and Chen, Zhan and Collister, Chua Jiahao and Huang, Qihang and Lv, Chen},
  journal={arXiv preprint arXiv:2512.01830},
  year={2025}
}

@article{tian2024drivevlm,
  title={Drivevlm: The convergence of autonomous driving and large vision-language models},
  author={Tian, Xiaoyu and Gu, Junru and Li, Bailin and Liu, Yicheng and Wang, Yang and Zhao, Zhiyong and Zhan, Kun and Jia, Peng and Lang, Xianpeng and Zhao, Hang},
  journal={arXiv preprint arXiv:2402.12289},
  year={2024}
}

@article{jiang2024senna,
  title={Senna: Bridging large vision-language models and end-to-end autonomous driving},
  author={Jiang, Bo and Chen, Shaoyu and Liao, Bencheng and Zhang, Xingyu and Yin, Wei and Zhang, Qian and Huang, Chang and Liu, Wenyu and Wang, Xinggang},
  journal={arXiv preprint arXiv:2410.22313},
  year={2024}
}

@article{zhang2024wisead,
  title={Wisead: Knowledge augmented end-to-end autonomous driving with vision-language model},
  author={Zhang, Songyan and Huang, Wenhui and Gao, Zihui and Chen, Hao and Lv, Chen},
  journal={arXiv preprint arXiv:2412.09951},
  year={2024}
}

@article{chi2025impromptu,
  title={Impromptu vla: Open weights and open data for driving vision-language-action models},
  author={Chi, Haohan and Gao, Huan-ang and Liu, Ziming and Liu, Jianing and Liu, Chenyu and Li, Jinwei and Yang, Kaisen and Yu, Yangcheng and Wang, Zeda and Li, Wenyi and others},
  journal={arXiv preprint arXiv:2505.23757},
  year={2025}
}

@inproceedings{xing2025openemma,
  title={Openemma: Open-source multimodal model for end-to-end autonomous driving},
  author={Xing, Shuo and Qian, Chengyuan and Wang, Yuping and Hua, Hongyuan and Tian, Kexin and Zhou, Yang and Tu, Zhengzhong},
  booktitle={Proceedings of the Winter Conference on Applications of Computer Vision},
  pages={1001--1009},
  year={2025}
}

@article{jiang2025diffvla,
  title={Diffvla: Vision-language guided diffusion planning for autonomous driving},
  author={Jiang, Anqing and Gao, Yu and Sun, Zhigang and Wang, Yiru and Wang, Jijun and Chai, Jinghao and Cao, Qian and Heng, Yuweng and Jiang, Hao and Dong, Yunda and others},
  journal={arXiv preprint arXiv:2505.19381},
  year={2025}
}

@article{gao2025diffvla++,
  title={DiffVLA++: Bridging Cognitive Reasoning and End-to-End Driving through Metric-Guided Alignment},
  author={Gao, Yu and Jiang, Anqing and Wang, Yiru and Jijun, Wang and Jiang, Hao and Sun, Zhigang and Yuwen, Heng and Shuo, Wang and Zhao, Hao and Hao, Sun},
  journal={arXiv preprint arXiv:2510.17148},
  year={2025}
}

@article{ang2026representational,
  title={From Representational Complementarity to Dual Systems: Synergizing VLM and Vision-Only Backbones for End-to-End Driving},
  author={Ang, Sining and Yang, Yuguang and Dang, Chenxu and Chen, Canyu and Chi, Cheng and Liu, Haiyan and Mao, Xuanyao and Bao, Jason and Sun, Bingchuan and Wang, Yan and others},
  journal={arXiv e-prints},
  pages={arXiv--2602},
  year={2026}
}

@article{wang2026hist,
  title={HiST-VLA: A Hierarchical Spatio-Temporal Vision-Language-Action Model for End-to-End Autonomous Driving},
  author={Wang, Yiru and Gu, Zichong and Gao, Yu and Jiang, Anqing and Sun, Zhigang and Wang, Shuo and Heng, Yuwen and Sun, Hao},
  journal={arXiv preprint arXiv:2602.13329},
  year={2026}
}

@article{li2026sgdrive,
  title={SGDrive: Scene-to-Goal Hierarchical World Cognition for Autonomous Driving},
  author={Li, Jingyu and Wu, Junjie and Hu, Dongnan and Huang, Xiangkai and Sun, Bin and Hao, Zhihui and Lang, Xianpeng and Zhu, Xiatian and Zhang, Li},
  journal={arXiv preprint arXiv:2601.05640},
  year={2026}
}

@article{liu2026driveworld,
  title={DriveWorld-VLA: Unified Latent-Space World Modeling with Vision-Language-Action for Autonomous Driving},
  author={Liu, Lin and Song, Ziying and Jia, Caiyan and Ye, Hangjun and Hao, Xiaoshuai and Chen, Long and others},
  journal={arXiv preprint arXiv:2602.06521},
  year={2026}
}

@article{schmidt2026lad,
  title={LAD-Drive: Bridging Language and Trajectory with Action-Aware Diffusion Transformers},
  author={Schmidt, Fabian and Fedurko, Karol and Enzweiler, Markus and Valada, Abhinav},
  journal={arXiv preprint arXiv:2603.02035},
  year={2026}
}

@article{zeng2025diffusionvl,
  title={DiffusionVL: Translating Any Autoregressive Models into Diffusion Vision Language Models},
  author={Zeng, Lunbin and Yao, Jingfeng and Liao, Bencheng and Tao, Hongyuan and Liu, Wenyu and Wang, Xinggang},
  journal={arXiv preprint arXiv:2512.15713},
  year={2025}
}

@article{zhu2025llada,
  title={Llada 1.5: Variance-reduced preference optimization for large language diffusion models},
  author={Zhu, Fengqi and Wang, Rongzhen and Nie, Shen and Zhang, Xiaolu and Wu, Chunwei and Hu, Jun and Zhou, Jun and Chen, Jianfei and Lin, Yankai and Wen, Ji-Rong and others},
  journal={arXiv preprint arXiv:2505.19223},
  year={2025}
}

@article{xu2025wam,
  title={WAM-Diff: A Masked Diffusion VLA Framework with MoE and Online Reinforcement Learning for Autonomous Driving},
  author={Xu, Mingwang and Cui, Jiahao and Cai, Feipeng and Shang, Hanlin and Zhu, Zhihao and Luan, Shan and Xu, Yifang and Zhang, Neng and Li, Yaoyi and Cai, Jia and others},
  journal={arXiv preprint arXiv:2512.11872},
  year={2025}
}

@article{dang2026drivefine,
  title={DriveFine: Refining-Augmented Masked Diffusion VLA for Precise and Robust Driving},
  author={Dang, Chenxu and Ang, Sining and Li, Yongkang and Tian, Haochen and Wang, Jie and Li, Guang and Ye, Hangjun and Ma, Jie and Chen, Long and Wang, Yan},
  journal={arXiv preprint arXiv:2602.14577},
  year={2026}
}

@article{ma2025dvlm,
  title={dVLM-AD: Enhance diffusion vision-language-model for driving via controllable reasoning},
  author={Ma, Yingzi and Cao, Yulong and Ding, Wenhao and Zhang, Shuibai and Wang, Yan and Ivanovic, Boris and Jiang, Ming and Pavone, Marco and Xiao, Chaowei},
  journal={arXiv preprint arXiv:2512.04459},
  year={2025}
}

@article{fu2025minddrive,
  title={MindDrive: A Vision-Language-Action Model for Autonomous Driving via Online Reinforcement Learning},
  author={Fu, Haoyu and Zhang, Diankun and Zhao, Zongchuang and Cui, Jianfeng and Xie, Hongwei and Wang, Bing and Chen, Guang and Liang, Dingkang and Bai, Xiang},
  journal={arXiv preprint arXiv:2512.13636},
  year={2025}
}

@article{yan2025ad,
  title={AD-R1: Closed-Loop Reinforcement Learning for End-to-End Autonomous Driving with Impartial World Models},
  author={Yan, Tianyi and Tang, Tao and Gui, Xingtai and Li, Yongkang and Zhesng, Jiasen and Huang, Weiyao and Kong, Lingdong and Han, Wencheng and Zhou, Xia and Zhang, Xueyang and others},
  journal={arXiv preprint arXiv:2511.20325},
  year={2025}
}

@article{rawal2026nord,
  title={NoRD: A Data-Efficient Vision-Language-Action Model that Drives without Reasoning},
  author={Rawal, Ishaan and Gupta, Shubh and Hu, Yihan and Zhan, Wei},
  journal={arXiv preprint arXiv:2602.21172},
  year={2026}
}

@article{luo2025adathinkdrive,
  title={Adathinkdrive: Adaptive thinking via reinforcement learning for autonomous driving},
  author={Luo, Yuechen and Li, Fang and Xu, Shaoqing and Lai, Zhiyi and Yang, Lei and Chen, Qimao and Luo, Ziang and Xie, Zixun and Jiang, Shengyin and Liu, Jiaxin and others},
  journal={arXiv preprint arXiv:2509.13769},
  year={2025}
}

@article{bai20243d,
  title={Is a 3d-tokenized llm the key to reliable autonomous driving?},
  author={Bai, Yifan and Wu, Dongming and Liu, Yingfei and Jia, Fan and Mao, Weixin and Zhang, Ziheng and Zhao, Yucheng and Shen, Jianbing and Wei, Xing and Wang, Tiancai and others},
  journal={arXiv preprint arXiv:2405.18361},
  year={2024}
}

@article{liu2025occvla,
  title={Occvla: Vision-language-action model with implicit 3d occupancy supervision},
  author={Liu, Ruixun and Kong, Lingyu and Li, Derun and Zhao, Hang},
  journal={arXiv preprint arXiv:2509.05578},
  year={2025}
}

@article{dang2026sparseoccvla,
  title={SparseOccVLA: Bridging Occupancy and Vision-Language Models via Sparse Queries for Unified 4D Scene Understanding and Planning},
  author={Dang, Chenxu and Wang, Jie and Li, Guang and Hou, Zhiwen and You, Zihan and Ye, Hangjun and Ma, Jie and Chen, Long and Wang, Yan},
  journal={arXiv preprint arXiv:2601.06474},
  year={2026}
}

@article{han2025percept,
  title={Percept-WAM: Perception-enhanced world-awareness-action model for robust end-to-end autonomous driving},
  author={Han, Jianhua and Tian, Meng and Zhu, Jiangtong and He, Fan and Zhang, Huixin and Guo, Sitong and Zhu, Dechang and Tang, Hao and Xu, Pei and Guo, Yuze and others},
  journal={arXiv preprint arXiv:2511.19221},
  year={2025}
}

@inproceedings{philion2020lift,
  title={Lift, splat, shoot: Encoding images from arbitrary camera rigs by implicitly unprojecting to 3d},
  author={Philion, Jonah and Fidler, Sanja},
  booktitle={European conference on computer vision},
  pages={194--210},
  year={2020},
  organization={Springer}
}

@inproceedings{li2023bevdepth,
  title={Bevdepth: Acquisition of reliable depth for multi-view 3d object detection},
  author={Li, Yinhao and Ge, Zheng and Yu, Guanyi and Yang, Jinrong and Wang, Zengran and Shi, Yukang and Sun, Jianjian and Li, Zeming},
  booktitle={Proceedings of the AAAI conference on artificial intelligence},
  volume={37},
  number={2},
  pages={1477--1485},
  year={2023}
}

@article{li2024bevformer,
  title={Bevformer: learning bird’s-eye-view representation from lidar-camera via spatiotemporal transformers},
  author={Li, Zhiqi and Wang, Wenhai and Li, Hongyang and Xie, Enze and Sima, Chonghao and Lu, Tong and Yu, Qiao and Dai, Jifeng},
  journal={IEEE Transactions on Pattern Analysis and Machine Intelligence},
  volume={47},
  number={3},
  pages={2020--2036},
  year={2024},
  publisher={IEEE}
}

@article{liang2022bevfusion,
  title={Bevfusion: A simple and robust lidar-camera fusion framework},
  author={Liang, Tingting and Xie, Hongwei and Yu, Kaicheng and Xia, Zhongyu and Lin, Zhiwei and Wang, Yongtao and Tang, Tao and Wang, Bing and Tang, Zhi},
  journal={Advances in neural information processing systems},
  volume={35},
  pages={10421--10434},
  year={2022}
}

@inproceedings{carion2020end,
  title={End-to-end object detection with transformers},
  author={Carion, Nicolas and Massa, Francisco and Synnaeve, Gabriel and Usunier, Nicolas and Kirillov, Alexander and Zagoruyko, Sergey},
  booktitle={European conference on computer vision},
  pages={213--229},
  year={2020},
  organization={Springer}
}

@inproceedings{wang2022detr3d,
  title={Detr3d: 3d object detection from multi-view images via 3d-to-2d queries},
  author={Wang, Yue and Guizilini, Vitor Campagnolo and Zhang, Tianyuan and Wang, Yilun and Zhao, Hang and Solomon, Justin},
  booktitle={Conference on robot learning},
  pages={180--191},
  year={2022},
  organization={PMLR}
}

@inproceedings{liu2022petr,
  title={Petr: Position embedding transformation for multi-view 3d object detection},
  author={Liu, Yingfei and Wang, Tiancai and Zhang, Xiangyu and Sun, Jian},
  booktitle={European conference on computer vision},
  pages={531--548},
  year={2022},
  organization={Springer}
}

@article{lin2022sparse4d,
  title={Sparse4d: Multi-view 3d object detection with sparse spatial-temporal fusion},
  author={Lin, Xuewu and Lin, Tianwei and Pei, Zixiang and Huang, Lichao and Su, Zhizhong},
  journal={arXiv preprint arXiv:2211.10581},
  year={2022}
}

@inproceedings{liu2023sparsebev,
  title={Sparsebev: High-performance sparse 3d object detection from multi-camera videos},
  author={Liu, Haisong and Teng, Yao and Lu, Tao and Wang, Haiguang and Wang, Limin},
  booktitle={Proceedings of the IEEE/CVF international conference on computer vision},
  pages={18580--18590},
  year={2023}
}

@article{Qwen3-VL,
      title={Qwen3-VL Technical Report}, 
      author={Shuai Bai and Yuxuan Cai and Ruizhe Chen and Keqin Chen and Xionghui Chen and Zesen Cheng and Lianghao Deng and Wei Ding and Chang Gao and Chunjiang Ge and Wenbin Ge and Zhifang Guo and Qidong Huang and Jie Huang and Fei Huang and Binyuan Hui and Shutong Jiang and Zhaohai Li and Mingsheng Li and Mei Li and Kaixin Li and Zicheng Lin and Junyang Lin and Xuejing Liu and Jiawei Liu and Chenglong Liu and Yang Liu and Dayiheng Liu and Shixuan Liu and Dunjie Lu and Ruilin Luo and Chenxu Lv and Rui Men and Lingchen Meng and Xuancheng Ren and Xingzhang Ren and Sibo Song and Yuchong Sun and Jun Tang and Jianhong Tu and Jianqiang Wan and Peng Wang and Pengfei Wang and Qiuyue Wang and Yuxuan Wang and Tianbao Xie and Yiheng Xu and Haiyang Xu and Jin Xu and Zhibo Yang and Mingkun Yang and Jianxin Yang and An Yang and Bowen Yu and Fei Zhang and Hang Zhang and Xi Zhang and Bo Zheng and Humen Zhong and Jingren Zhou and Fan Zhou and Jing Zhou and Yuanzhi Zhu and Ke Zhu},
	  journal={arXiv preprint arXiv:2511.21631},
      year={2025}
}

@article{tschannen2025siglip,
  title={Siglip 2: Multilingual vision-language encoders with improved semantic understanding, localization, and dense features},
  author={Tschannen, Michael and Gritsenko, Alexey and Wang, Xiao and Naeem, Muhammad Ferjad and Alabdulmohsin, Ibrahim and Parthasarathy, Nikhil and Evans, Talfan and Beyer, Lucas and Xia, Ye and Mustafa, Basil and others},
  journal={arXiv preprint arXiv:2502.14786},
  year={2025}
}

@article{jia2024bench2drive,
  title={Bench2drive: Towards multi-ability benchmarking of closed-loop end-to-end autonomous driving},
  author={Jia, Xiaosong and Yang, Zhenjie and Li, Qifeng and Zhang, Zhiyuan and Yan, Junchi},
  journal={Advances in Neural Information Processing Systems},
  volume={37},
  pages={819--844},
  year={2024}
}

@inproceedings{xie2025vlms,
  title={Are VLMs Ready for Autonomous Driving? An Empirical Study from the Reliability, Data and Metric Perspectives},
  author={Xie, Shaoyuan and Kong, Lingdong and Dong, Yuhao and Sima, Chonghao and Zhang, Wenwei and Chen, Qi Alfred and Liu, Ziwei and Pan, Liang},
  booktitle={Proceedings of the IEEE/CVF International Conference on Computer Vision},
  pages={6585--6597},
  year={2025}
}

@article{liao2022maptr,
  title={Maptr: Structured modeling and learning for online vectorized hd map construction},
  author={Liao, Bencheng and Chen, Shaoyu and Wang, Xinggang and Cheng, Tianheng and Zhang, Qian and Liu, Wenyu and Huang, Chang},
  journal={arXiv preprint arXiv:2208.14437},
  year={2022}
}

@inproceedings{gu2023vip3d,
  title={Vip3d: End-to-end visual trajectory prediction via 3d agent queries},
  author={Gu, Junru and Hu, Chenxu and Zhang, Tianyuan and Chen, Xuanyao and Wang, Yilun and Wang, Yue and Zhao, Hang},
  booktitle={Proceedings of the IEEE/CVF Conference on Computer Vision and Pattern Recognition},
  pages={5496--5506},
  year={2023}
}

@article{chai2019multipath,
  title={Multipath: Multiple probabilistic anchor trajectory hypotheses for behavior prediction},
  author={Chai, Yuning and Sapp, Benjamin and Bansal, Mayank and Anguelov, Dragomir},
  journal={arXiv preprint arXiv:1910.05449},
  year={2019}
}

@article{wang2026vggdrive,
  title={VGGDrive: Empowering Vision-Language Models with Cross-View Geometric Grounding for Autonomous Driving},
  author={Wang, Jie and Li, Guang and Huang, Zhijian and Dang, Chenxu and Ye, Hangjun and Han, Yahong and Chen, Long},
  journal={arXiv preprint arXiv:2602.20794},
  year={2026}
}

@article{zhu2025sparsead,
  title={Sparsead: Sparse query-centric paradigm for efficient end-to-end autonomous driving},
  author={Zhu, Runwen and Zhao, Jianbo and Zhang, Diankun and Wang, Guoan and Chen, Xiwu and Zhang, Siyu and Gong, Jiahao and Zhou, Qibin and Zhang, Wenyuan and Wang, Ningzi and others},
  journal={IEEE Transactions on Artificial Intelligence},
  year={2025},
  publisher={IEEE}
}

@article{shang2026dynvla,
  title={DynVLA: Learning World Dynamics for Action Reasoning in Autonomous Driving},
  author={Shang, Shuyao and Zhan, Bing and Yan, Yunfei and Wang, Yuqi and Li, Yingyan and An, Yasong and Wang, Xiaoman and Liu, Jierui and Hou, Lu and Fan, Lue and others},
  journal={arXiv preprint arXiv:2603.11041},
  year={2026}
}

@article{luo2026last,
  title={LaST-VLA: Thinking in Latent Spatio-Temporal Space for Vision-Language-Action in Autonomous Driving},
  author={Luo, Yuechen and Li, Fang and Xu, Shaoqing and Ji, Yang and Zhang, Zehan and Wang, Bing and Shen, Yuannan and Cui, Jianwei and Chen, Long and Chen, Guang and others},
  journal={arXiv preprint arXiv:2603.01928},
  year={2026}
}

@article{luo2026unleashing,
  title={Unleashing VLA Potentials in Autonomous Driving via Explicit Learning from Failures},
  author={Luo, Yuechen and Chen, Qimao and Li, Fang and Xu, Shaoqing and Liu, Jaxin and Song, Ziying and Yang, Zhi-xin and Wen, Fuxi},
  journal={arXiv preprint arXiv:2603.01063},
  year={2026}
}

@article{mao2023gpt,
  title={Gpt-driver: Learning to drive with gpt},
  author={Mao, Jiageng and Qian, Yuxi and Ye, Junjie and Zhao, Hang and Wang, Yue},
  journal={arXiv preprint arXiv:2310.01415},
  year={2023}
}

@article{hu2022lora,
  title={Lora: Low-rank adaptation of large language models.},
  author={Hu, Edward J and Shen, Yelong and Wallis, Phillip and Allen-Zhu, Zeyuan and Li, Yuanzhi and Wang, Shean and Wang, Liang and Chen, Weizhu and others},
  journal={Iclr},
  volume={1},
  number={2},
  pages={3},
  year={2022}
}

@article{wiedmann2025finevision,
  title={Finevision: Open data is all you need},
  author={Wiedmann, Luis and Zohar, Orr and Mahla, Amir and Wang, Xiaohan and Li, Rui and Frere, Thibaud and von Werra, Leandro and Gosthipaty, Aritra Roy and Marafioti, Andr{\'e}s},
  journal={arXiv preprint arXiv:2510.17269},
  year={2025}
}

@article{loshchilov2017decoupled,
  title={Decoupled weight decay regularization},
  author={Loshchilov, Ilya and Hutter, Frank},
  journal={arXiv preprint arXiv:1711.05101},
  year={2017}
}

@article{chen2024we,
  title={Are we on the right way for evaluating large vision-language models?},
  author={Chen, Lin and Li, Jinsong and Dong, Xiaoyi and Zhang, Pan and Zang, Yuhang and Chen, Zehui and Duan, Haodong and Wang, Jiaqi and Qiao, Yu and Lin, Dahua and others},
  journal={Advances in Neural Information Processing Systems},
  volume={37},
  pages={27056--27087},
  year={2024}
}

@inproceedings{yue2024mmmu,
  title={Mmmu: A massive multi-discipline multimodal understanding and reasoning benchmark for expert agi},
  author={Yue, Xiang and Ni, Yuansheng and Zhang, Kai and Zheng, Tianyu and Liu, Ruoqi and Zhang, Ge and Stevens, Samuel and Jiang, Dongfu and Ren, Weiming and Sun, Yuxuan and others},
  booktitle={Proceedings of the IEEE/CVF conference on computer vision and pattern recognition},
  pages={9556--9567},
  year={2024}
}

@article{fu2023mme,
  title={Mme: A comprehensive evaluation benchmark for multimodal large language models},
  author={Fu, Chaoyou and Chen, Peixian and Shen, Yunhang and Qin, Yulei and Zhang, Mengdan and Lin, Xu and Yang, Jinrui and Zheng, Xiawu and Li, Ke and Sun, Xing and others},
  journal={arXiv preprint arXiv:2306.13394},
  year={2023}
}

@inproceedings{rahmanzadehgervi2024vision,
  title={Vision language models are blind},
  author={Rahmanzadehgervi, Pooyan and Bolton, Logan and Taesiri, Mohammad Reza and Nguyen, Anh Totti},
  booktitle={Proceedings of the Asian Conference on Computer Vision},
  pages={18--34},
  year={2024}
}

@inproceedings{masry2022chartqa,
  title={Chartqa: A benchmark for question answering about charts with visual and logical reasoning},
  author={Masry, Ahmed and Do, Xuan Long and Tan, Jia Qing and Joty, Shafiq and Hoque, Enamul},
  booktitle={Findings of the association for computational linguistics: ACL 2022},
  pages={2263--2279},
  year={2022}
}

@article{yan2025m,
  title={M 2 diffuser: Diffusion-based trajectory optimization for mobile manipulation in 3d scenes},
  author={Yan, Sixu and Zhang, Zeyu and Han, Muzhi and Wang, Zaijin and Xie, Qi and Li, Zhitian and Li, Zhehan and Liu, Hangxin and Wang, Xinggang and Zhu, Song-Chun},
  journal={IEEE Transactions on Pattern Analysis and Machine Intelligence},
  year={2025},
  publisher={IEEE}
}

@article{huang2026automot,
  title={AutoMoT: A Unified Vision-Language-Action Model with Asynchronous Mixture-of-Transformers for End-to-End Autonomous Driving},
  author={Huang, Wenhui and Zhang, Songyan and Huang, Qihang and Wang, Zhidong and Mao, Zhiqi and Chua, Collister and Chen, Zhan and Chen, Long and Lv, Chen},
  journal={arXiv preprint arXiv:2603.14851},
  year={2026}
}

@article{zuo2026vega,
  title={Vega: Learning to Drive with Natural Language Instructions},
  author={Zuo, Sicheng and Li, Yuxuan and Zheng, Wenzhao and Zhu, Zheng and Zhou, Jie and Lu, Jiwen},
  journal={arXiv preprint arXiv:2603.25741},
  year={2026}
}

@article{marouani2026revisiting,
  title={Revisiting [CLS] and Patch Token Interaction in Vision Transformers},
  author={Marouani, Alexis and Sim{\'e}oni, Oriane and J{\'e}gou, Herv{\'e} and Bojanowski, Piotr and Vo, Huy V},
  journal={arXiv preprint arXiv:2602.08626},
  year={2026}
}
}

\end{document}